\documentclass[10pt,twocolumn,letterpaper]{article}

\usepackage{cvpr}              % To produce the CAMERA-READY version
\usepackage[dvipsnames]{xcolor}

\DeclareMathAlphabet{\mathmybb}{U}{bbold}{m}{n}
\usepackage{color, colortbl}
\definecolor{LightCyan}{rgb}{0.94,0.94,0.94}
\usepackage{arydshln}
\usepackage{multirow}
\usepackage{makecell}

\usepackage{algorithm}
\usepackage{algorithmic}
\usepackage{newfloat}
\usepackage{listings}
\floatstyle{ruled}
\newfloat{listing}{tb}{lst}{}
\floatname{listing}{Listing}

\definecolor{cvprblue}{rgb}{0.21,0.49,0.74}
\usepackage[pagebackref,breaklinks,colorlinks,citecolor=cvprblue]{hyperref}

\def\paperID{1583} % *** Enter the Paper ID here
\def\confName{CVPR}
\def\confYear{2024}

\title{MVFormer: Diversifying Feature Normalization and Token Mixing \\for Efficient Vision Transformers}

\author{Jongseong Bae$^{1}$\thanks{Equally contributed} \hspace{1em}
Susang Kim$^{1}$\footnotemark[1] \hspace{1em}
Minsu Cho$^{2}$ \hspace{1em}
Ha Young Kim$^{1}$\thanks{Corresponding author}\\
$^{1}$Yonsei University\hspace{1em}
$^{2}$Pohang University of Science and Technology
\\
\tt{\small{js.bae@yonsei.ac.kr \hspace{1em} healess@yonsei.ac.kr }}\\ \tt{\small{ mscho@postech.ac.kr \hspace{1em} hayoung.kim@yonsei.ac.kr}}
}
\begin{document}
\maketitle

\begin{abstract}
\label{abstract}
Active research is currently underway to enhance the efficiency of vision transformers (ViTs).
Most studies have focused solely on effective token mixers, overlooking the potential relationship with normalization.
To boost diverse feature learning, we propose two components: a normalization module called multi-view normalization (MVN) and a token mixer called multi-view token mixer (MVTM).
The MVN integrates three differently normalized features via batch, layer, and instance normalization using a learnable weighted sum.
Each normalization method outputs a different distribution, generating distinct features.
Thus, the MVN is expected to offer diverse pattern information to the token mixer, resulting in beneficial synergy.
The MVTM is a convolution-based multiscale token mixer with local, intermediate, and global filters, and it incorporates stage specificity by configuring various receptive fields for the token mixer at each stage, efficiently capturing ranges of visual patterns.
We propose a novel ViT model, multi-vision transformer (MVFormer), adopting the MVN and MVTM in the MetaFormer block, the generalized ViT scheme.
Our MVFormer outperforms state-of-the-art convolution-based ViTs on image classification,
object detection, and instance and semantic segmentation with the same or lower parameters and MACs.
Particularly, MVFormer variants, MVFormer-T, S, and B achieve 83.4\%, 84.3\%, and 84.6\% top-1 accuracy, respectively, on ImageNet-1K benchmark.
\end{abstract}

\section{Introduction}
\label{sec:intro}
Vision transformers (ViTs)  have achieved great success in the computer vision field~\cite{dosovitskiy2020image}.
\begin{figure}[ht]
  \centering
  \includegraphics[width=\linewidth]
  {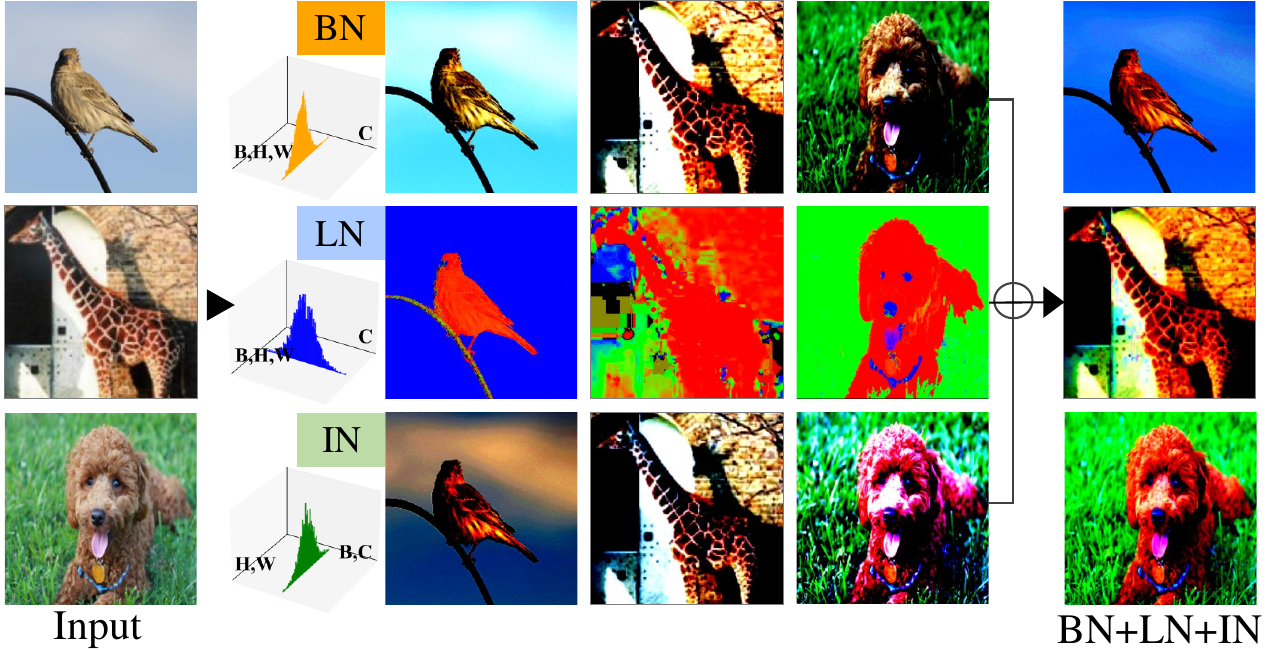}
  % \begin{tabular}{m{14mm}m{12mm}m{13mm}m{10mm}m{14mm}}
  %      (a) Input&(b) BN &(c) LN &(d) IN &(e) BN+LN+IN
  % \end{tabular}
  \caption{%Visualization of normalized images in image and fourier spectrum using BN, LN, and their averages. 
  \textbf{Visualization of normalized images from BN, LN, IN, and their averages.} These illustrate that BN and IN maintain the detailed spatial distribution of the input image, whereas LN overly smooths the image. We can intuitively observe the spatially smoothed output, including the local details, by taking a simple average.  
  % By taking a simple average, we can intuitively observe the spatially smoothed output to include the local details. 
  % The second row shows that the distinction between them is also able to be observed in frequency domain.
  }
  \label{fig:norm_vis}
\vspace*{-.2cm}
\end{figure}
Since self-attention in traditional transformers~\cite{vaswani2017attention} has been in the spotlight, numerous studies have proposed various effective and efficient spatial mixing methods, referred to as token mixers, to improve or substitute self-attention. 
% in this work, 
Some studies~\cite{chu2021twins,fan2021multiscale,yang2021focal,li2022mvitv2,Wang_2022,kim2021relational} have proposed attention-variant methods, such as Swin~\cite{liu2021swin}, to enhance the efficiency of traditional self-attention,
% efficiency of token mixers, 
whereas others~\cite{tolstikhin2021mlp,yang2022focal,wang2022shift, chen2021cyclemlp} have proposed competitive non-attention token mixers.
Among the currently available options, the convolutional operator has recently been applied in the transformer block. 
For example, the ConvNeXt~\cite{liu2022convnet} model is a milestone connecting convolution with ViT that modernizes the convolutional neural network (CNN)
by introducing a transformer variant scheme.
Recent studies have demonstrated the benefit of appropriate inductive bias in ViT~\cite{park2022vision,raghu2021vision}, which has emerged as an ongoing research topic~\cite{yang2022focal,dai2021coatnet,tu2022maxvit, guo2022segnext, yu2023inceptionnext}.

MetaFormer~\cite{yu2022metaformer,yu2022metaformer1} is an abstracted architecture scheme derived from the transformer, in which the token mixer is not specified. 
While token mixers have been the primary focus to ensure feature diversity, the other components in recent ViTs have generally been based on MetaFormer~\cite{wang2023riformer}.
Among these components, we concentrate on normalization.
% which is able to provide various feature distributions 
% to the token mixer.
% , allowing it to adaptively select and use them.
Batch normalization (BN)~\cite{ioffe2015batch}, layer normalization (LN)~\cite{ba2016layer}, and instance normalization (IN)~\cite{ulyanov2017instance} produce distinct distributions and different features due to their varying normalizing dimensions.
Inspired by this, we conduct a simple visualization to observe the changes that occur when the differently normalized images are integrated, as illustrated in Fig.~\ref{fig:norm_vis}.
Each method emphasizes specific patterns in the input image. All these patterns are also visible in the composite image, which is an average of the three normalized images. 
%Through this observation, we expect the token mixer to become capable of learning a more distinct set of features provided by various normalization distributions.
Through this observation, we confirmed that integrating various normalizations can convey a diverse set of features with various distributions to the token mixer.
In this work, we introduce a normalization module, multi-view normalization (MVN), to diversify feature learning.
% which adaptively combines various characteristics of common normalization techniques.
%MVN simply weights and sums three differently normalized features via BN, LN and instance normalization (IN)~\cite{ulyanov2017instance}, widely used method in style transfer task~\cite{7780634}, in order to alleviate the batch-dependency of BN.
The MVN combines three differently normalized features via BN, LN, and IN, using a learnable weighted sum.
% IN is widely used in style transfer task~\cite{7780634}, which is included in MVN to alleviate the mini-batch dependency of BN.
In this manner, MVN can flexibly reflect the diverse specificities, such as batch-level, channel-level, and sample-level dependencies, providing various feature distributions to the token mixer and enabling it to use them adaptively.
The experiments confirm that this simple mechanism significantly improves the performance with a negligible increase in parameters and computational costs.
%In addition, MVN even shows notable generalization performances in other existing ViTs such as ConvNexT and Poolformer.
Moreover, MVN can be easily applied to existing ViTs and CNNs, such as Swin~\cite{liu2021swin} and ConvNeXt~\cite{liu2022convnet}, consistently improving their original performances.
On top of that, experimental results strongly support the insight that the unique attributes of each normalization play meaningful roles in performance, and their appropriate combination creates beneficial synergy.

In addition, to diversify the mixing range of token mixers further, we propose a convolutional token mixer, called the multi-view token mixer (MVTM).
Similar to the latest convolution-based ViTs~\cite{guo2022segnext,yu2023inceptionnext}, the MVTM is a multiscale depthwise convolutional operator that employs multiple mixing filters channelwise with different receptive fields. 
Beyond the existing paradigm of bisecting local and global mixing filters, the MVTM consists of local, intermediate, and global mixing filters to enhance its mixing capacity.
% Additionally, to efficientize stage-level multi-scale token mixing,
% the MVTM introduces the stage-specificity that differently constitutes the volume of each-level mixing filter depending on the preferred range of the receptive field at each stage~\cite{park2022vision, yu2022metaformer, yu2022metaformer1}.
Furthermore, the MVTM introduces stage specificity, which varies the volume of each level of mixing filter and global mixing filter size differently depending on the preferred range of the receptive field at each stage to make stage-level multiscale token mixing efficient~\cite{park2022vision, yu2022metaformer, yu2022metaformer1}.
% to efficientize stage-level multi-scale token mixing, the MVTM introduces the stage-specificity that differently varies the volume of each-level mixing filter and global mixing filter size depending on the preferred range of the receptive field at each stage~\cite{park2022vision, yu2022metaformer, yu2022metaformer1}.

% Adopting our MVN and MVTM into the MetaFormer block, 
We propose a novel convolution-based ViT, the multi-vision transformer (MVFormer), by adopting the MVN and MVTM in the MetaFormer block. The MVFormer addresses the existing demands of token mixers to capture diverse patterns from multiple perspectives, further extending it to normalization.
Boosted by the enhanced capacity of the viewpoint, the MVFormer displays notable efficiency and effectiveness through extensive experiments.
The MVFormer outperforms other existing convolution-based ViTs
% Our MVFormer achieves performance matching that of existing other convolution-based ViTs.
on image classification and downstream tasks, object detection, and instance and semantic segmentation,
%on ImageNet-1K classification task,
with equal or even fewer parameters and multiply-accumulates (MACs).
%, object detection and semantic segmentation, 
% even with lower parameters and computational costs.
Particularly, MVFormer-tiny (-T), small (S), and base (B), variants of the MVFormer, achieve state-of-the-art (SOTA) performance of 83.4\%, 84.3\%, and 84.6\%, respectively, on the ImageNet-1K benchmark~\cite{5206848}. 
% Specially in the case of MVFormer-S, it even shows competitive performance compared to those of other -B models, reducing more than 40\% of parameters and FLOPs.
The main contributions of this work are summarized as follows:
\begin{itemize}
\item 
%This work initially proposes a normalization integration paradigm that combines differently normalized features, especially purposing to elaborate convolution-based ViT.
%We introduce a novel normalization module, Multi-View Normalization, which simply integrates three normalized features via BN, LN and IN. %IN is for mitigating batch-dependency of BN.
%Our MVN adaptively reflects the unique specifities of each normalization method.
%We introduce Multi-View Normalization, MVN, which adaptively integrates three normalized features via BN, LN and IN with learnable weigths. This study is the first to introduce a normalization integration paradigm in ViTs.
%We propose Multi-View Normalization (MVN), integrating various normalized features, for diversifying feature learning, and it is the first instance of a normalization integration paradigm in ViTs.
We propose MVN, which integrates various normalized features to diversify feature learning, providing a range of feature distributions to the token mixer. It is the first study of a normalization integration paradigm in ViTs.
The MVN significantly enhances performance with negligible increases in parameters and computational costs. 
% In addition, MVN is even applicable to existing ViTs and CNNs with notable performance enhancements.

% in transformer block to increase feature diversity with mixed three normalized features through BN, LN and IN. Our extensive experiments support that this simple method significantly promotes the model performance with no additional parameter and ignorable FLOPs.
% we investigate the compatibility between normalization method and the receptive field of token mixing method in vision transformer block. We identify the Fourier spectra of batch normalization and layer normalization, then compare with those of CNN kernels with various receptive field.

\item  
% MVFormer utilizes three-scale depthwise convolution consisting of local, mid-level and global mixing filters.
%We propose a multi-scale convolutional token mixer, Multi-View Token Mixier, composed of local, mid-level and global mixing filters to effectively capture diverse spatial patterns.
%We propose the Multi-View Token Mixier, a multi-scale convolutional token mixer composed of local, mid-level and global mixing filters, to effectively capture diverse spatial patterns.
We introduce MVTM, a multiscale convolutional token mixer, to better capture diverse spatial patterns.
%MVTM also reflects stage-specificity, differently setting the receptive field of token mixer depending on the preferred mixing scale at each stage, which effectively exploits feature pyramid structure.
The MVTM also reflects stage specificity, setting the receptive field of the token mixer differently based on the preferred mixing scale at each stage, effectively exploiting the feature pyramid structure.

\item %As adopting MVN and MVTM into the MetaFormer block, we propose a novel convolution-based ViT, MVFormer. 
%As adopting MVN and MVTM into the MetaFormer block, we present MVFormer. %Our MVFormer-variants, MVFormer-T, -S and -B outperform existing convolution based ViTs strong performance with lower parameters and FLOPs on performances of 82.5\%, 83.8\%
%and 83.9\% on ImageNet-1K classification, in order.
%Our MVFormer-variants, MVFormer-T, -S and -B outperform existing convolution based ViTs with lower parameters and FLOPs on performances of 82.5\%, 83.8\%
%and 83.9\% on ImageNet-1K classification, in order.
By adopting MVN and MVTM in the MetaFormer block, we present MVFormer, surpassing the existing convolution-based ViTs on image classification, object detection, and instance and semantic segmentation with the same or even fewer parameters and MACs.
% classification%, object detection and semantic segmentation.
% Especially, our MVFormer-T, MVFormer-S and MVFormer-B achieve
 % strong performance with lower parameters and FLOPs on performances of 82.5\%, 83.8\%
% and 83.9\% on ImageNet-1K classification, in order.
\end{itemize}

\section{Related Work}
\label{sec:related work}
\subsection{Normalization for Computer Vision Tasks}
%Normalization approach has attracted major interest in the literature as an essential component in deep neural networks such as CNN, ViT. But normalization is widely recognized for its ability to enhance the speed  and stability of training processes.
%Normalization approaches have been studied as an essential component in deep neural networks such as CNNs and ViTs, especially in computer vision, due to their ability to enhance the speed and stability of training.
% Normalization approaches have been studied as an essential component in deep neural networks such as CNNs and ViTs due to their ability to enhance the speed and stability of training.
Normalization methods have been investigated as a key component of deep neural networks to enhance the training speed and stability. Typically, BN~\cite{ioffe2015batch} plays a pivotal role in CNNs for vision-related tasks.
However, its minibatch dependency has been demonstrated to cause performance degradation with a small batch size on several vision tasks, such as semantic segmentation~\cite{ba2016layer}.
To improve this, several BN variants, such as batch renormalization~\cite{ioffe2017batch}, EvalNorm~\cite{singh2019evalnorm}, MABN~\cite{yan2020stabilizing}, and PowerNorm~\cite{shen2020powernorm} have been proposed.
The LN~\cite{ba2016layer} first emerged in natural language processing (NLP) to address the summed input in recurrent neural networks.
Compared to BN, LN calculates the channel statistics equally on all data points. 
As LN was adopted in the initial transformer, it has been employed in recent ViTs~\cite{dosovitskiy2020image,touvron2021training,liu2021swin}.
Group normalization (GN) is a generalized LN that calculates grouped channel statistics.
A previous study proposed modified LN (MLN), which equates GN with a single group, to improve the performance of PoolFormer~\cite{yu2022metaformer}.
% Related to this, Poolformer~\cite{yu2022metaformer} suggestes modified layer normalization (MLN), which equals GN with a single group.
Additionally, a study has investigated the use of BN for ViTs~\cite{9607565},
% There also exists the study to leverage BN for ViTs~\cite{9607565}, 
% as inserting its parameters into the linear layer.
by inserting its parameters into the linear layer.
% ViT is very appropriate for Layer Normalization (LN)~\cite{ba2016layer} that only calculates statistics in the channel dimension while ignoring the batch and sequence length dimensions. %##
% Meanwhile, 
IN~\cite{ulyanov2017instance} is widely used in style transfer, such as in AdaIN~\cite{huang2017arbitrary}, which is a representative IN-variant to transplant the style information of input features.
Moreover, spatially modulated normalization techniques, such as SPADE~\cite{park2019semantic} and MoTo~\cite{qian2022makes}, were proposed to prevent information loss, and global response normalization (GRN)~\cite{woo2023convnext} aims to enhance the interchannel feature diversity.
% Instance Normalization(IN) can be altered by normalizing the feature statistics. 
% It has been observed that the style information of an image is normalized in the pixel space, which can be normalized by considering each channel. ##
In contrast to these studies, we propose an initial paradigm of combining the existing normalization approaches in ViTs.
% However, it is possible to combine these techniques from different points of view in order to get enhanced characteristics.

% \subsection{CNN Based Vision Model}
% Since 2012, when AlexNet~\cite{krizhevsky2012imagenet} succeeded in ILSVRC 2012, CNN has been predominant for vision backbone architecture. 
% VGG~\cite{simonyan2014very}, GoogLeNet~\cite{szegedy2014going}, ResNet~\cite{he2016deep} and DenseNet~\cite{huang2018densely} progressively succeeded in stacking deeper network.
% ResNeXt~\cite{xie2017aggregated} combined the philosophies of Inception in GoogLeNet and ResNet to enhance the model capacity.
% MobileNet~\cite{howard2017mobilenets}, ShuffleNet~\cite{zhang2017shufflenet} and EfficientNet~\cite{tan2020efficientnet} designed efficient network architecture as still maintaining competitive performance compared to that of previous models. 
% Traditionally, CNN backbone models utilize BN in common, in order to accelerate and stabilize model training.https://www.overleaf.com/project/64a0b36bd1fd1b1f200c2801

\subsection{Vision Transformer with Token Mixer}
%\subsection{Attention-Based Token Mixer}
% Motivated by 
The success of the transformer in NLP led to its use in computer vision.
% The success of Transformers in NLP has motivated, the need for vision Transformer. 
Previous studies have reported the impressive performance of
ViT~\cite{dosovitskiy2020image} and DeiT~\cite{touvron2021training} on image classification, and extended the application of the Swin transformer~\cite{liu2021swin} to object detection and semantic segmentation. However, owing to the high computational cost of self-attention, studies have attempted to replace it with alternative token mixers.
% there have been studies aiming to replace it with alternative token mixers.
% there have arisen studies to substitute it with other token mixers. 
Accordingly, multilayer perceptron (MLP)-like token mixers~\cite{tolstikhin2021mlp,touvron2021resmlp,lian2021mlp,yu2022s2,wei2022activemlp} have emerged as a dominant approach that employs an MLP operator to mix spatial tokens.
% belong to one dominant approach that employs MLP operator to mix spatial tokens.
% Additionally, studies have investigated the use of convolutional network as the token mixer. For example, a previous study proposed the use of ConvNeXt~\cite{liu2022convnet}, modernized paradigm of CNN, completely substitute traditional self-attention in transformers with depthwise convolution.
As another mainstream approach, depthwise convolution has been studied as a token mixer. The ConvNeXt~\cite{liu2022convnet} model applies the modernized paradigm of the CNN, completely substituting traditional self-attention in the transformer with depthwise convolution. 
Further, other studied models, such as FocalNet~\cite{yang2022focal} and VAN~\cite{guo2022visual}, employ convolution-based attention mechanisms, enabling the model to capture input-dependent token interactions.
% suggested convolution-based attention mechanisms, enabling the model to capture input-dependent token interactions.
% Further, another study proposed FocalNet~\cite{yang2022focal}, a convolution-based Focal Modulation, which introduces the gate mechanism to model input-dependent interactions.
The ConvFormer~\cite{yu2022metaformer1} model is the current SOTA convolution-based ViT, introducing the inverted separable convolution in MobileNetV2~\cite{sandler2019mobilenetv2} as the token mixer.
Recently, multiscale convolutional token mixers~\cite{guo2022segnext,yu2023inceptionnext} have been introduced that employ multiple mixing paths parallelly to reflect local and global information effectively.
This study applies an advanced multiscale depthwise convolution, employing an intermediate mixing filter and the concept of stage specificity.
% that parallely employs multiple mixing paths to effectively reflect both local and global informations.
% Further on this, in this work, we present advanced multi-scale depthwise convolution that incorporates a newly introduced mid-level mixing filter. Additionally, we adopt the concept of stage-specificity to the token mixer to reflect the feature pyramid structure of the convolution-based ViTs.
% Existing convolution based ViTs have unified block structure across all stages.
% In this work, we further inject stage-level inductive bias 
% But convolution frequently encounter a limitation in their expressivity due to the fixed size of its kernels, they do not take into account the decreasing size of the feature map at each stage.

% Due to the heavy computational cost required by self-attention, following studies such as Twins~\cite{chu2021twins}, MViT~\cite{fan2021multiscale}, Focal Transformer~\cite{yang2021focal}, MViT v2~\cite{li2022mvitv2} and PvT v2~\cite{Wang_2022} have struggled to mitigate this while maintaining or even outperforming SoTA performance. Different from CNN models, LN has been consistently used in Transformer architecture including original model for NLP. 
% VAN~\cite{guo2022visual} adopts the decomposed convolution module, which is a sequence of depthwise convolution, dilated convolution and pointwise convolution layers.
\begin{figure*}[t]
  \centering
  \includegraphics[width=\linewidth]{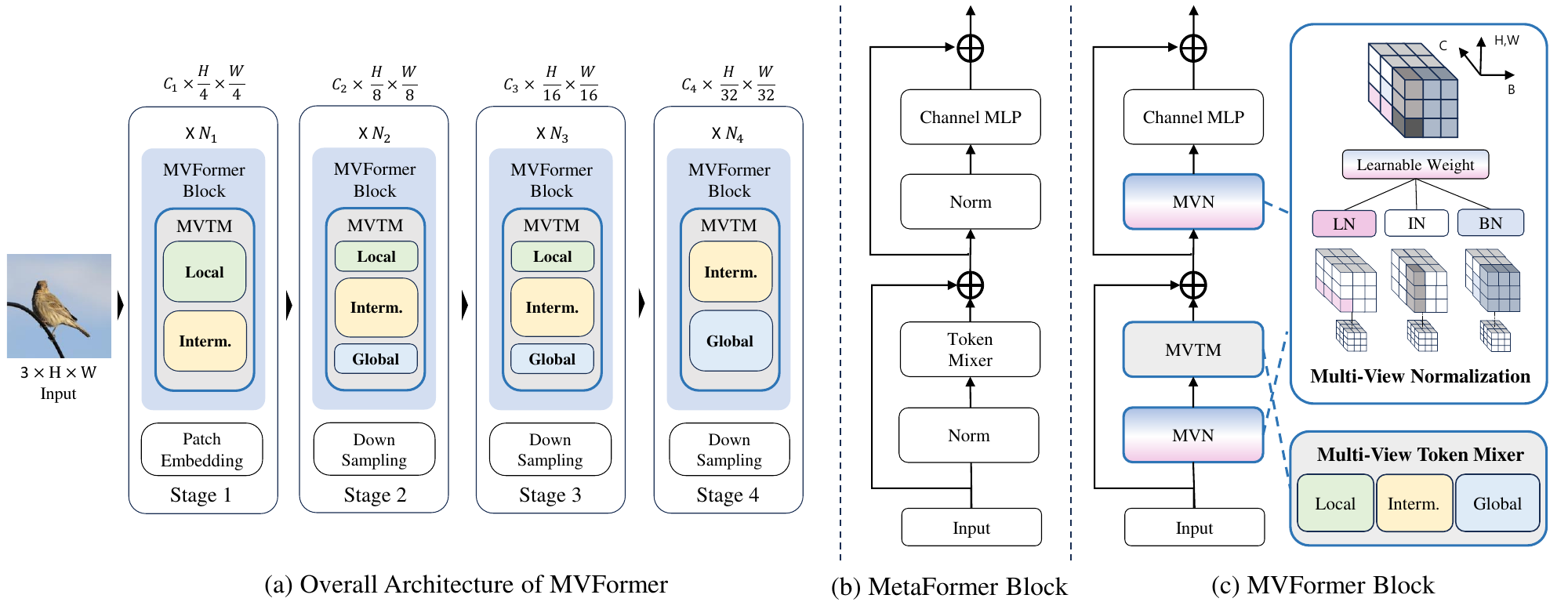}
  \caption{\textbf{Overall architecture of the proposed MVFormer and MVFormer block.} Similar to MetaFormer, each block of MVFormer adopts a hierarchical architecture with four stages. Each $\mathrm{Stage}_j$ comprises $N_j$ blocks with a feature dimension $C_j$. The MVFormer block consists of two main components, MVN and MVTM, which can be compared to the MetaFormer block.
  }% TriNo is a channel-wise fusion of three normalized BN, LN, and IN features. To extract multi-scale features, we utilize MSTM, which consists of filters operating at the local, mid, and global levels.
  \label{fig:overall_architecture}
\vspace*{-.2cm}
\end{figure*}
\section{Method}
\subsection{Preliminaries}
\subsubsection{MetaFormer}
MetaFormer~\cite{yu2022metaformer,yu2022metaformer1} is an abstracted general architecture of modern ViTs as presented in Fig.~\ref{fig:overall_architecture}~(b).
A batch of input images, $I \in \mathbb{R}^{B\times C_{I}\times H\times W}$, can be patchified into the patch embedding, $X \in \mathbb{R}^{B\times C\times \frac{H}{P}\times \frac{W}{P}}$, where $B, H$ and $W$ denote the batch size, height, and width of $I$, respectively, $C_{I}$ and $C$ represent the embedding dimension of $I$ and $X$, respectively.
In addition, $P$ corresponds to the patch size.
The embedded feature $X$ passes repeated MetaFormer blocks, and the output of each block, $Y\in \mathbb{R}^{B\times C\times \frac{H}{P}\times \frac{W}{P}}$, is calculated as follows:
\begin{align}
\hat{X} &= \mathrm{Token Mixer}(\mathrm{Norm}_1(X))+X,
\label{eq:tokenmixer}\\
Y &= \mathrm{MLP}(\mathrm{Norm}_2(\hat{X}))+\hat{X}. 
\label{eq:mlp_norm}
\end{align}

% MetaFormer block consists of $Token Mixer$ and $MLP$ parts, and each part includes the normalization and residual connection, respectively.
The MetaFormer block consists of a token mixer subblock ($\mathrm{Token Mixer}(\cdot)$) and an MLP subblock ($\mathrm{MLP}(\cdot)$). Each subblock includes normalization ($\mathrm{Norm}_1(\cdot)$ and $\mathrm{Norm}_2(\cdot)$) and a residual connection.
%, respectively.
% MetaFormer block consists of two sub-blocks, which are a token mixer sub-block ($Token Mixer$) and a MLP sub-block ($MLP$), and each sub-block includes the normalization ($Norm_1$ and $Norm_2$) and residual connection, respectively.
The $\mathrm{Token Mixer}$ is not specified, corresponding to certain spatial mixing modules, such as self-attention or convolution, and $\mathrm{MLP}$ denotes the two-layer feed-forward network with an activation function.
% like ReLU~\cite{agarap2019deep} or GELU~\cite{hendrycks2023gaussian}. However, we employed StarReLU~\cite{yu2022metaformer1} based on the MetaFormer.
% For both $Norm_1$ and $Norm_2$, LN has been commonly utilized. 
This work follows the overall paradigm of MetaFormer, with ConvFormer~\cite{yu2022metaformer1} as the baseline.
% introducing the token mixer of ConvFormer~\cite{yu2022metaformer1} as the baseline.
% incorporating the token mixer from ConvFormer~\cite{yu2022metaformer1} as our baseline.
%introducing the token mixer and normalization components based on our proposed approach from ConvFormer~\cite{yu2022metaformer1} as the baseline.

% Whole architecture is composed of four stages.
% When shifting to next stage, the spatial shape of feature map is downsized in half, while the embedding dimension is doubled.
% Finally, the output feature is input to the additional heads for downstream tasks such as image classification, object detection and semantic segmentation.

\subsubsection{Normalization}
The BN~\cite{ioffe2015batch}, LN~\cite{ba2016layer}, and IN~\cite{ulyanov2017instance} are commonly used in vision architectures.
Both BN and LN were proposed to accelerate model training, and IN was introduced for the image stylization method.
Although these methods similarly normalize the feature distribution, their normalizing dimensions differ.
% Those similarly normalize feature distribution, however, the normalizing dimensions of three methods are all different.
Given an input feature $X  \in \mathbb{R}^{B\times C\times H\times W}$, the output of each method is calculated as follows:
\begin{align}
X_{BN} &= \frac{X - \mu_{B,H,W}}{\sigma_{B,H,W}},\\
X_{LN} &= \frac{X - \mu_{C}}{\sigma_{C}},\\
X_{IN} &= \frac{X - \mu_{H,W}}{\sigma_{H,W}},
\end{align}
where $\mu$ and $\sigma$ are the mean and standard deviation of $X$, which is calculated based on the normalization dimensions indicated by the subscripts.
%As mentioned in Introduction section, BN channel-wisely normalizes feature distribution, on the other hand, LN is pixel-wisely applied. 
% As mentioned in Introduction section, 
The BN normalizes the feature distribution channelwise, whereas LN operates at the pixel level. 
%IN is similar to BN, which normalizes sample-level spatial distribution.
Moreover, IN is similar to BN, as it normalizes sample-level spatial distributions. A channelwise affine transform commonly follows each output.

\subsection{Multi-Vision Transformer}
This section, explains the details of the MVFormer.
The overview of MVFormer is presented in Fig.~\ref{fig:overall_architecture}~(a).

\subsubsection{Multi-View Normalization}
% Since the common normalization techniques such as BN, LN and IN similarly normalize the input features, they have been considered as substitutable options in a network, although they are strictly distinguished in normalizing dimension that is an important factor changing the output distribution.
Common normalization techniques, such as BN, LN, and IN, similarly normalize the input features; thus, they are considered substitutable options in a network. However, these techniques are distinguished by their normalizing dimensions, a crucial factor changing the output distribution.
% For example, 
% % while BN and LN work on channels and pixels separately, 
% while each of BN and LN operates channel- and pixel-wisely, BN changes the channel distribution of input feature whereas LN changes the spatial distribution of each channel.
% In addition, as comparing BN and IN, IN is differentiated by its mini-batch independent characteristic.
We expect this distribution variation to influence the overall feature learning to extract visual patterns. From the perspective of feature diversity, the model can explore the extended manifold, where all varied distributions are provided.
% We observed that the visual pattern of the input feature changes depending on the normalization methods.
% To diversify the input feature while preserving its identity,
% In the perspective of feature diversity, 
% % rather than speed-up of model training, 
% we observed that the visual pattern of the input feature changes depending on the normalization methods.
%Aiming to efficiently train ViT exposed on various visual feature patterns, we initially suggest the normalization integration paradigm.
% Thus, we initially suggest to employ the normalization integration paradigm for the efficient training of ViT on diverse visual feature patterns.
% Hence, we initially propose adopting the normalization integration paradigm to efficiently train ViT on a variety of visual feature patterns.
Therefore, we propose a novel normalization integration paradigm to enhance performance by training the ViT on a diverse range of feature distributions.
% Employing feature diversity is advantageous to improving representation learning. So 
% There are numerous techniques to integrate multiple features. 
% Among them, we select a simple leanable weighted summation which is efficient and capable of adaptively combining multiple features without significantly changing the original distributions of each individual feature.%which is efficient, and capable to adaptively combine multiple features while not much chaning original distributions of each feature.
% which is efficient and capable of adaptively combining multiple features without significantly changing the original distributions of each individual feature.
% MVN is to maintain all dimension-wise input feature distributions partially kept in each normalized features.
We designed a normalization module, MVN, which uses a learnable weighted summation of three normalized features obtained through BN, LN, and IN.
%Through this mechanism, MVN enables the model to simultaneously capture unique specificities in each normalized feature.
Through this mechanism, MVN allows the model to capture unique specificities of each normalized feature simultaneously, enabling it to pass on more diverse features to the token mixer.

%Mathematically
The input feature $X\in \mathbb{R}^{B\times C\times H\times W}$ is first transformed into $X_{BN}, X_{LN}$, and $X_{IN}$, respectively, then summed with learnable weights.
The resulting output feature $X_{MVN}\in \mathbb{R}^{B\times C\times H\times W}$, which is calculated as follows:
\begin{align}
% \centering
X_{MVN} = \alpha_{BN}X_{BN}+\alpha_{LN}X_{LN}+\alpha_{IN}X_{IN},
% \lambda_{BN} + \lambda_{LN} + \lambda_{IN} = 1,
% Y_{BN} &= Token Mixer_{BN}(BN(X_{BN}))+X_{BN},\\
% Y_{LN} &= Token Mixer_{LN}(LN(X_{LN}))+X_{LN},\\https://www.overleaf.com/project/64a0b36bd1fd1b1f200c2801
% Y &= Concat(Y_{BN},Y_{LN}),
% Z &= MLP(Norm(Y))+Y,
\end{align}
where $\alpha_{BN}$, $\alpha_{LN}$, and $\alpha_{IN} \in \mathbb{R}^{C}$ are learnable parameters whose dimension sizes are equal to the embedding dimension of $X$.
% There are numerous techniques to integrate multiple features. 
% Among them, we select a simple leanable weighted summation which is efficient and capable of adaptively combining multiple features without significantly changing the original distributions of each individual feature.
% To maintain the overall scale of input feature, we restrict the summed weights to be 1 at all channel through taking softmax fucntion ($\alpha_{BN}$ + $\alpha_{LN}$ + $\alpha_{IN}$ = $\mathmybb{1}^C$).
% In this manner, MVN enables the model to flexibly learn appropriate combination of three normalized feature.
%In this manner, MVN enables the model to flexibly search preferred combination of various normalized features, just with slight amount of additional parameters and FLOPs.
% For precise training of $\alpha_{BN}$, $\alpha_{LN}$, and $\alpha_{IN}$, 
% the affine transform is applied to $X_{MVN}$ at once, instead applied to each normalization method.
% For adaptive training of features generated by 
To enable the model to search for the precise ratio of $\alpha_{BN}$, $\alpha_{LN}$, and $\alpha_{IN}$, 
the affine transform is applied to $X_{MVN}$ at once instead of to each normalization method.
In this manner, the MVN can flexibly explore the preferred combination of various normalized features with just a slight number of additional parameters and MACs.

% We select weighted sum which is more efficent, and 

\subsubsection{Multi-View Token Mixer}
% Recent convolution-based ViTs adopt multi-scale depthwise convolution that channel-wisely diversifies the kernel sizes to inject various spatial inductive biases~\cite{guo2022segnext, yu2023inceptionnext}. 
% In computer vision, designing multi-scale networks~\cite{szegedy2016rethinking,guo2022segnext,yu2023inceptionnext} is a widespread direction to obtain multi-scale features. 
% @@? convonlution-based
Recent studies on convonlution-based ViTs have presented notable performance. These studies used multiscale depthwise convolution~\cite{guo2022segnext, yu2023inceptionnext}, which diversifies its kernel size channelwise to add various spatial inductive biases. 
% However, these studies did not take into account how the receptive field changes with each stage, so we focus on this aspect.
In practice, these studies have primarily been based on a dichotomous perspective of distinguishing only local and global mixing.
% Different from the attention mechanism, convolution is static method that the learned filters slide in data-independent manner, thus the diversification of receptive field is required to extract various range of visual patterns.
In contrast to the attention mechanism, which dynamically adjusts weights based on input values, convolution operates as a static method, where filters slide in a data-independent manner.
% In contrast to the attention mechanism, which dynamically adjusts weights based on input values, convolution is a static method that a learnable filter slide across entire feature in a data-independent manner.
Consequently, the receptive field must be diversified to extract a wide range of visual patterns.
% beyond this, we aim to approach this mechanism in extended viewpoint of feature diversity.
From this viewpoint, we propose a three-scale convolutional token mixer, MVTM, consisting of local, global, and newly added intermediate mixing filters to capture the intermediate range of visual patterns between the local and global receptive fields.
% By this, we expect to elaborate the robustness of convolutional token mixer on visual object scale. 
We expect this approach to mitigate the heterogeneity between local and global mixed features, and we elaborate on the robustness of the convolutional token mixer on the visual object scale.

% extends the model capacity as boosting the diverse feature learning.
% Foremost, we adopt the token mixer of ConvFormer~\cite{yu2022metaformer1}, which is current SOTA convolution-based ViT, as the baseline.
% Foremost, we adopts the inverted separable convolution module in MobileNetV2~\cite{sandler2019mobilenetv2}, utilized as the token mixer in ConvFormer~\cite{yu2022metaformer1}.

The MVTM is based on the inverted separable convolutional module of MobileNetV2~\cite{sandler2019mobilenetv2}, which is used as the token mixer in ConvFormer~\cite{yu2022metaformer1}.
% adopt the token mixer of ConvFormer~\cite{yu2022metaformer1}, the current SOTA convolution-based ViT, as our baseline.
%, specifying these token mixers as commonly-used operators to probe the model's potential for achieving state-of-the-art performance. 
% It utilizes the inverted separable convolution module in MobileNetV2~\cite{sandler2019mobilenetv2}, 
For a normalized feature $X_N\in \mathbb{R}^{B\times C\times H\times W}$, the output feature $\Tilde{X}\in \mathbb{R}^{B\times C\times H\times W}$ is calculated as follows: 
\begin{align}
\Tilde{X} = \mathrm{PWConv}_2(\mathrm{DWConv}(\sigma(\mathrm{PWConv}_1(X_N)))),
\label{eq:convformer}
\end{align}
where $\mathrm{DWConv}$ and $\mathrm{PWConv}$ denote the depthwise and pointwise convolutional layers, respectively, and $\sigma$ indicates the activation function. % $PWConv_1$ and $PWConv_2$ respectively upsizes and downsizes the channel dimension twice.
Unlike the baseline that uses a 7$\times$7 depthwise convolutional layer for $\mathrm{DWConv}$, MVTM diversifies the receptive field of $\mathrm{DWConv}$ as follows:
% Based on this mechanism, we focus to elaborate $DWConv$, which is the spatial mixing layer.
% @@ ~ that this approach enables mitigate heterogeneity ~
% ConvNeXt\cite{liu2022convnet} conducted a number of experiments to find the optimal kernel size, and their findings indicated that a kernel size of 7x7 achieved the best outcomes. Therefore, the mid-level kernel was defined as 7x7.
% Our MVTM operates as described in Equation~\ref{eq:convformer}, in which $DWConv$ layer is specified as follows:
% Given an input feature $X_{in}\in \mathbb{R}^{B\times C\times H\times W}$, our MVTM mixes visual tokens as follows:
\begin{align}
{X}^\mathrm{l}_{\mathrm{out}} &= \mathrm{DWConv}_\mathrm{l}(X_{\mathrm{in}}^\mathrm{l}), \label{eq:dwc_l}\\
{X}^\mathrm{i}_{\mathrm{out}} &= \mathrm{DWConv}_\mathrm{m}(X_{\mathrm{in}}^\mathrm{i}), \label{eq:dwc_i}\\
{X}^\mathrm{g}_{\mathrm{out}} &= \mathrm{DWConv}_\mathrm{g}(X_{\mathrm{in}}^\mathrm{g}), \label{eq:dwc_g}\\
{X}_{\mathrm{out}} &= \mathrm{Concat}({X}^\mathrm{l}_{\mathrm{out}},{X}^\mathrm{i}_{\mathrm{out}},{X}^\mathrm{g}_{\mathrm{out}}). \label{eq:dwc_concat}
\end{align}
%where $\hat{X}_{l}$, $\hat{X}_{i}$ and $\hat{X}_{g}$ channel-wisely split from $\hat{X}$ for local, intermediate and global mixing, respectively.

First, $X_{\mathrm{in}}\in \mathbb{R}^{B\times C\times H\times W}$ is split into three channel-wise groups of $X_{\mathrm{in}}^\mathrm{l}\in \mathbb{R}^{B\times C_\mathrm{l}\times H\times W}$, $X_{\mathrm{in}}^\mathrm{i}\in \mathbb{R}^{B\times C_\mathrm{i}\times H\times W}$, and $X_{\mathrm{in}}^\mathrm{g}\in \mathbb{R}^{B\times C_\mathrm{g}\times H\times W}$. Each of these groups denotes the channel-split features from $X_{\mathrm{in}}$ for local, intermediate, and global mixing, where $C_\mathrm{l}, C_\mathrm{i}$, and $C_\mathrm{g}$ denote their corresponding channels, respectively ($C_\mathrm{l} + C_\mathrm{i} + C_\mathrm{g} = C$).
% channel-wise split features from $X_N$ 
% local, mid-level and global for spatial mixing, respectively.
Those are individually input into $\mathrm{DWConv}_\mathrm{l}$, $\mathrm{DWConv}_\mathrm{i}$, and $\mathrm{DWConv}_\mathrm{g}$, which are token mixing depthwise convolutional layers with local, intermediate, and global kernel sizes, respectively; the kernel sizes in $\mathrm{DWConv}_\mathrm{l}$ and $\mathrm{DWConv}_\mathrm{i}$ are fixed as 3$\times$3 and 7$\times$7, respectively. As for $\mathrm{DWConv}_\mathrm{g}$, its kernel size is adjusted at each stage, as explained in the next paragraph.
% to rearrange its receptive field fitting on varying input shape at each stage.
% whose kernel sizes are all different.
% and the subscripts "l", "m", and "g" in $DWConv$ represent the local, mid-level, and global mixing filters, respectively.
%%whose kernel sizes are all different; those in $DWConv_l$ and $DWConv_m$ are fixed as 3$\times$3 and 7$\times$7, respectively, and for $DWConv_g$, it is adjusted at each stage, as explaining in next paragraph.
% Each mixed features are channel-wisely concatenated, then finally fused by $Fusion$ module, which employs a 1$\times$1 pointwise convolution.
Finally, each mixed feature is concatenated channelwise.
This mechanism endows the MVTM with the capability to capture multiple ranges of visual representations.
% @@ endows or provides / ranges
% fixed as 3$\times$3 and 7$\times$7 for $DWConv_l$ and $DWConv_m$, respectively,
% and in case of $DWConv_g$, it is differently set at each stage, since the global shape of feature varies with stage number.

%Additionally, MVTM further introduces the concept of stage-specificity.
Furthermore, MVTM introduces the concept of stage specificity.
Recent ViTs primarily follow the feature pyramid structure that systemically downsizes the feature shape at the beginning of each stage~\cite{liu2021swin,liu2022convnet,yu2022metaformer1,fan2021multiscale}.
According to the previous studies on ViT architecture~\cite{park2022vision, yu2022metaformer}, it has been analytically and experimentally observed that 
% there exists the preferred range of token mixing scale at each stage.
% For instance, 
employing local constraints on a token mixer is effective in the initial stages. In contrast, % In specific, local constraints on token mixer is effective in early stages, on the other hand, 
wide mixing for global token interaction is required in the late stages.
This property has not been adopted in convolution-based ViTs, as a fixed kernel design is applied across all token mixing layers.

For the first time, we adopted this paradigm to enable convolution-based ViT to capture various ranges of visual patterns efficiently.
% @@ initially <-> for the first time ?
% it is required to specify local and global receptive field at each stage
% Since the convolution 
% To apply this into the convolutional token mixer, it is required to specify 
To implement this, we regulated two configurations of MVTM: 1) a channelwise ratio of three mixing filters and 2) the kernel size of the global mixing filter ($\mathrm{DWConv}_\mathrm{g}$).
The former is to determine the predominant mixing scale of MVTM, and the latter is for rearranging the scope of global mixing with varying input shapes.
% In this manner, we aim to enable MVTM to efficiently extract various range of visual information in the feature pyramid structure.
In this manner, MVTM weights the preferred receptive field at each stage depending on the input.
% incorporating filters designed to reflect the characteristics of a pyramid structure, we aim to extract a diverse range of feature information.
Table~\ref{tab:method_MSTM} details the configurations.
% The detailed configurations are specified in Table~\ref{tab:method_MSTM}.
% @@ purpose /= verb, need 'the': In this manner, we aim to maximize the impact of MVTM in the feature pyramid structure.
\begin{table}[hbt!]
	\centering
    \setlength\tabcolsep{6pt}\renewcommand{\arraystretch}{.9}
      \resizebox{\linewidth}{!}{
    \begin{tabular}{c|c|c|c|c}
        \Xhline{1.0pt}
        Configuration & Stage 1 & Stage 2 & Stage 3 & Stage 4  \\
        \hline
        $C_\mathrm{l}:C_\mathrm{i}:C_\mathrm{g} (\%)$ & 50:50:0 & 25:50:25 & 25:50:25 & 0:50:50   \\
        % Adaptive &  $7\times7$  &  $7\times7$  &  $7\times7$  &  $7\times7$ \\
        \Xhline{0.1pt}
        \multirow{2}{*}{Global kernel size} & $55\times 1$ &  $27\times1$  &  $13\times1$  &  \multirow{2}{*}{$7\times7$}   \\
       &$1\times 55$ &  $1\times27$  &  $1\times13$  &  \\
        \Xhline{1.0pt}
    \end{tabular}}
    \caption{\textbf{Stage specific configurations of MVTM.} For global kernel sizes in Stages 1, 2, and 3, we decompose $\mathrm{DWConv}_\mathrm{g}$ into horizontal and vertical mixing kernels for efficiency. 
    % channel-wisely parallelize horizontal and vertical mixing kernels for computational efficiency.
    }
    \vspace{-.2cm}
    \label{tab:method_MSTM}
\end{table}  

%We design MVTM in stage-specific manner as increasing the channel-wise ratio of local to global filters, and conversely, decreasing the global mixing filter size step by step depending on the stage number.
In MVTM, as the stage number increases, we increase the channelwise ratio of local to global filters and decrease the size of the global mixing filter step by step. We expect the MVTM to captures a productive range of visual information efficiently at each stage.
% In the case of global mixing kernel size, it systemically reduces in half to fit its receptive field to the downsized input size from the previous stage.
% We specifically show the effectiveness of these mechanisms on convolution-based ViT in Experiment section.

% $n \in \left\{55,27,13,7\right\}$ decreases as the stage goes deeper. 

\subsubsection{MVFormer Block} 
% As introducing our MVN and MVTM into the MetaFormer block,
As introducing MVN and MVTM into the MetaFormer block, we propose the MVFormer block, as presented in Fig.~\ref{fig:overall_architecture}~(c).
In the MVFormer block, MVN first extracts various feature distributions. Based on this, MVTM explores diversified feature spaces for token mixing.
In addition, by equally inserting MVN in the MLP subblock, we expect a particular beneficial interaction between them, similar to that in the token mixer subblock.
% we expect certain interaction between diverse visual features from MVN and the ability to capture various range of token connectivities in MVTM.
% In this structure, we expect certain interaction between diverse visual features from MVN and the ability to capture various range of token connectivities in MVTM.
We reformulated Eq.~\ref{eq:tokenmixer} and Eq.~\ref{eq:mlp_norm} as follows:
% \begin{align}
% &\hat{Z} = Concat(LN(Y_{BN}),LN(Y_{LN})), \label{eq:concat_norm}\\
% &Z = MLP(\hat{Z})+Y.
% \end{align}
\begin{align}
\hat{X} &= \mathrm{MVTM}(\mathrm{MVN}(X))+X, 
\label{eq:tokenmixer_ours}\\
Y &= \mathrm{MLP}(\mathrm{MVN}(\hat{X}))+\hat{X},
\label{eq:mlp_norm_ours}
\end{align}
% where $\mathrm{MVN}$ and $\mathrm{MVTM}$ refer our MVN and MVTM.
% We apply our MVN to both $Norm_1$ and $Norm_2$, and adopt MVTM as the token mixer in the MetaFormer block as shown in Figure~\ref{fig:overall_architecture}~(c).
%For the $MLP$ module, which is the same as in Eq.~\ref{eq:mlp_norm}, using GELU activation function.
where the $\mathrm{MLP}$ module is the same as in Eq.~\ref{eq:mlp_norm}. 
For the activation functions in $\mathrm{MVTM}$ and $\mathrm{MLP}$, StarReLU~\cite{yu2022metaformer1} is used.
% We name this MetaFormer-variant block as the tri-normalized block (TriNo block).
% $Y_{BN}$ and $Y_{LN}$, because both features are heterogeneous as each of them is independently processed.
% Those are finally mixed through $MLP$ module, which is the same as in Eq.\ref{eq:mlp_norm}.

\subsubsection{Overall Architecture}
% Combining the benefits of three normalizations and incorporating stage-specific properties
%in Figure~\ref{fig:overall_architecture}~(c).
Considering the unique specificities of the three normalized features and multiple scale-mixed features with stage specificity, we propose an effective convolution-based ViT, MVFormer. 
The overall paradigm of MVFormer is the same with MetaFormer when introducing MVN and MVTM into the MetaFormer blocks.
Depending on the  parameters and computational complexity, MVFormer is categorized into MVFormer-xT, MVFormer-T, MVFormer-S, and MVFormer-B, where MVFormer-xT is the primary model for feasibility.
The specific configurations of each MVFormer model are described in the Appendix~\ref{model_config}. 
\begin{table}[hbt!]
	\centering
    \setlength\tabcolsep{2.5pt}\resizebox{\linewidth}{!}{
    	\begin{tabular}{l|c|c|cc|c}
    	    \Xhline{1.0pt}
    		\multirow{2}{*}{Method}&  \multirow{2}{*}{Type}&\multirow{2}{*}{Input} & \#Params & MACs & Top-1 acc. \\
    		 ~ & ~ && (M) & (G) & (\%)  \\
           	\Xhline{1.0pt}

       %          \multicolumn{6}{c}{Tiny Models}  \\
       %          %                 \hline 
       %          % ResMLP-S12  & MLP &$224^2$& 15 & 3.0 & 76.6 \\
       %          % CycleMLP-B1 & MLP &$224^2$& 15 & 2.1 & 78.9 \\
       %          % ATMNet-xT  & MLP&$224^2$ & 15 & 2.2 & 79.7 \\
       %                          \hline 

       %          ResNet18 & Conv&$224^2$ & 12 & 1.8 & 69.8 \\            FasterNet-T2 & Conv&$224^2$ & 15 & 1.9 & 78.9 \\
       %          \rowcolor{LightCyan}
    		 % \textbf{MVFormer-xT  } &   \textbf{Conv}&$224^2$ &   \textbf{17} &   \textbf{2.2} & \textbf{81.3}\\   
       %                 \hline 

                       \multicolumn{6}{c}{Small Models}  \\

                \hline 
                DeiT-S~\cite{touvron2021training} & Attn &$224^2$& 22 & 4.6 & 79.8 \\
                % T2T-ViT-14 & Attn &$224^2$& 22 & 4.8 & 81.5 \\
                % TNT-S & Attn& $224^2$& 24 & 5.2 & 81.5 \\
                PVT-Small~\cite{wang2021pyramid} & Attn &$224^2$& 25 & 3.8 & 79.8 \\

                Swin-T~\cite{liu2021swin} & Attn &$224^2$& 29 & 4.5 & 81.3 \\
                Focal-T~\cite{yang2021focal} & Attn& $224^2$& 29 & 4.9 & 82.2 \\
                % \hline 

                % CycleMLP-B2 & MLP&$224^2$ & 27 & 3.9 & 81.6 \\
                % MorphMLP-T  & MLP &$224^2$& 23 & 3.9 & 81.6 \\
                % ActiveMLP-T  & MLP &$224^2$& 27 & 4.0 & 82.0 \\

                % PoolFormer-S36 & MLP& $224^2$& 31 & 5.0 & 81.4 \\

                \hline 
                % RSB-ResNet-50 & Conv &$224^2$& 26 & 4.1 & 79.8 \\
                % RegNetY-4G~\cite{radosavovic2020designing}  & Conv&$224^2$ & 21 & 4.0 & 81.3 \\
                ConvNeXt-T~\cite{liu2022convnet}  & Conv &$224^2$& 29 & 4.5 & 82.1 \\
                FocalNet-T~\cite{yang2022focal} & Conv&$224^2$ & 29 & 4.5 & 82.3 \\
                InceptionNeXt-T~\cite{yu2023inceptionnext}& Conv& $224^2$& 28 & 4.2 & 82.3 \\
                VAN-B2~\cite{guo2022visual}& Conv& $224^2$& 27 & 5.0 & 82.8 \\
                ConvFormer-S18~\cite{yu2022metaformer1}& Conv& $224^2$& 27 & 3.9 & 83.0 \\

                \rowcolor{LightCyan}
                \textbf{MVFormer-T  } &   Conv&$224^2$ &   \textbf{27} &   \textbf{3.9} & \textbf{83.4}\\   
                \hdashline

                ConvFormer-S18~\cite{yu2022metaformer1}& Conv& $384^2$& 27 & 11.6 & 84.4 \\

                \rowcolor{LightCyan}
                \textbf{MVFormer-T  } &   Conv&$384^2$ &   \textbf{27} &   \textbf{11.5} & \textbf{84.8}\\   
                \hline 
                
                      \multicolumn{6}{c}
                       {Medium Models}  \\
                \hline 
                % PVT-Medium~\cite{wang2021pyramid} & Attn &$224^2$& 44 & 6.7 & 81.2 \\
                PVT-Large~\cite{wang2021pyramid}& Attn&$224^2$ & 61 & 9.8 & 81.7 \\
                Swin-S~\cite{liu2021swin} & Attn &$224^2$& 50 & 8.7 & 83.0 \\
                Focal-S~\cite{yang2021focal} & Attn&$224^2$ & 51 & 9.1 & 83.5 \\
                
                % \hline 

                % CycleMLP-B4 & MLP&$224^2$ & 50 & 10.1 & 83.0\\
                %  MorphMLP-B  & MLP &$224^2$& 58 & 10.2 & 83.2 \\

                % ActiveMLP-B  & MLP &$224^2$& 52 & 10.1 & 83.5 \\

                % PoolFormer-M36 & MLP&$224^2$ & 56 & 8.8 & 82.1 \\

                \hline 
                % RSB-ResNet-101 & Conv&$224^2$ & 45 & 7.9 & 81.3 \\
                % RegNetY-8G~\cite{radosavovic2020designing}  & Conv &$224^2$& 39 & 8.0 & 82.1 \\
                ConvNeXt-S~\cite{liu2022convnet}   & Conv&$224^2$ & 50 & 8.7 & 83.1 \\
                FocalNet-S~\cite{yang2022focal} & Conv& $224^2$& 50 & 8.7 & 83.5 \\
                InceptionNeXt-S~\cite{yu2023inceptionnext} & Conv &$224^2$& 49 & 8.4 & 83.5 \\
                VAN-B3~\cite{guo2022visual}& Conv& $224^2$& 45 & 9.0 & 83.9 \\
                ConvFormer-S36~\cite{yu2022metaformer1}& Conv& $224^2$& 40 & 7.6 & 84.1 \\

                \rowcolor{LightCyan}
    		 \textbf{MVFormer-S  } &   Conv &$224^2$&   \textbf{40} &   \textbf{7.6} & \textbf{84.3}\\     
                \hdashline
                ConvFormer-S36~\cite{yu2022metaformer1}& Conv& $384^2$& 40 & 22.4 & 85.4 \\

                \rowcolor{LightCyan}
                \textbf{MVFormer-S  } &   Conv&$384^2$ &   \textbf{40} &   \textbf{22.2} & \textbf{85.6}\\   
                \hline  
                                       \multicolumn{6}{c}{Large Models}  \\
                \hline 
                DeiT-B~\cite{touvron2021training} & Attn&$224^2$ & 86 & 17.5 & 81.8 \\
                % T2T-ViT-14 & Attn&$224^2$ & 64 & 13.8 & 82.3 \\
                % TNT-B & Attn&$224^2$ & 66 & 14.1 & 82.9 \\
                Swin-B~\cite{liu2021swin} & Attn &$224^2$& 88 & 15.4 & 83.5 \\
                Focal-B~\cite{yang2021focal} & Attn&$224^2$ & 90 & 16.0 & 83.8 \\
                % \hline 
                % CycleMLP-B5 & MLP& $224^2$& 76 & 12.3 & 83.2 \\
                % MorphMLP-L  & MLP &$224^2$& 76 & 12.5 & 83.4 \\
                % ActiveMLP-L  & MLP &$224^2$& 76 & 12.3 & 83.6 \\
                % % PoolFormer-M48 & MLP&$224^2$ & 73 & 11.6 & 82.5 \\
                \hline 
                % RSB-ResNet-152 & Conv& $224^2$& 60 & 11.6 & 81.8 \\
                % RegNetY-16G~\cite{radosavovic2020designing}   & Conv&$224^2$ & 84 & 15.9 & 82.2 \\
                ConvNeXt-B~\cite{liu2022convnet}   & Conv &$224^2$& 89 & 15.4 & 83.8 \\
                FocalNet-B~\cite{yang2022focal}& Conv&$224^2$ & 89 & 15.4 & 83.9 \\
                InceptionNeXt-B~\cite{yu2023inceptionnext} & Conv &$224^2$& 87 & 14.9 & 84.0 \\
                VAN-B4~\cite{guo2022visual}& Conv& $224^2$& 60 & 12.2 & 84.2 \\
                ConvFormer-M36~\cite{yu2022metaformer1}& Conv& $224^2$& 57 & 12.8 & 84.5 \\
                
                \rowcolor{LightCyan}
    		 \textbf{MVFormer-B  } &   Conv&$224^2$ &   \textbf{57} &   \textbf{12.7} & {\textbf{84.6}}\\ 
                \hdashline
                
                ConvNeXt-B~\cite{liu2022convnet} & Conv& $384^2$& 89 & 45.0 & 85.1 \\
                InceptionNeXt-B~\cite{yu2023inceptionnext}& Conv& $384^2$& 87 & 43.6 & 85.2 \\
                ConvFormer-M36~\cite{yu2022metaformer1}& Conv& $384^2$& 57 & 37.7 & 85.6 \\

                \rowcolor{LightCyan}
                \textbf{MVFormer-B  } &   Conv&$384^2$ &   \textbf{57} &   \textbf{37.4} & \textbf{85.7}\\   
       %  \hline
       % \multicolumn{6}{c}
                %        {Medium Models$\uparrow$}  \\
                % \hline 
                % ConvNext-S  & Conv&$224^2$ & 50 & 22.5 & 83.1 \\
                % \textbf{MVFormer-S } &   \textbf{Conv} &384&   \textbf{46} &   \textbf{24.5} & \textbf{}\\
                % \hline
                % \multicolumn{6}{c}
                %        {Large Models$\uparrow$}\\
                %     \hline
                %     ConvNext-B  & Conv &384& 89 & 45.0 & 85.1 \\
                %     InceptionNeXt-B & Conv &384& 87 & 43.6 & 85.2 \\
                %     \textbf{MVFormer-B } &   \textbf{Conv}&384 &   \textbf{80} &   \textbf{43.3} & {\underbar{}}\\
            \Xhline{1.0pt}
            \end{tabular}
}
    
    \caption{\textbf{Comparison with SOTA models on ImageNet-1K.} }
    % All these models are only trained on the ImageNet1K training set, and the validation set accuracy is reported.}
    \label{tab:sota_in1k}
\vspace*{-.2cm}
\end{table}  
\section{Experiments}
%To validate the effectiveness of the MVFormer on visual representation learning, we performed the proposed MVFormer in terms of normalization and kernel size related on the feature map size. This is to be compared with static kernel size. We evaluate the performance of MVFormer on three mainstream visual recognition benchmarks: image classifcation on ImageNet-1kdataset\cite{deng2009imagenet}, object detection on COCO dataset\cite{lin2014microsoft} and semantic segmentation on ADE20k dataset\cite{zhou2019semantic}. Furthermore, we visualize the experimental results and fourier spetrum on ImageNet validation set.
\subsection{Image Classification}
We conduct image classification experiments on the ImageNet-1K benchmark~\cite{5206848}, including 1.28M training images and 50K validation images from 1K classes.
To augment and regularize the input images for training, we employ  weight decay, %random clipping, random horizontal flipping,
RandAugment~\cite{cubuk2020randaugment}, Random Erasing~\cite{zhong2020random}, Mixup~\cite{zhang2017mixup}, CutMix~\cite{yun2019cutmix}, Label Smoothing~\cite{szegedy2016rethinking}, Stochastic Depth~\cite{huang2016deep} and training strategy of DeiT~\cite{touvron2021training}. We train all models from scratch for 300 epochs with an input size of 224$\times$224.
We use the AdamW~\cite{kingma2014adam,loshchilov2017decoupled} optimizer with a cosine learning rate schedule, including 20 warm-up epochs. ResScale~\cite{shleifer2021normformer} is used for the last two stages. The batch size, learning rate, and weight decay are set to 4096, 4e-3, and 0.05, respectively. 
We also use the stochastic depth with a probability of 0.2 for MVFormer-xT and MVFormer-T and 0.3 and 0.4 for MVFormer-S and MVFormer-B, respectively.
% We also used stochastic depth~\cite{huang2016deep} with the probability of 0.2 for MVFormer-xT and MVFormer-T, and 0.3 and 0.4 for MVFormer-S, MVFormer-B, respectively.
%We also use stochastic depth with the probability of 0.2 for MVFormer-xT and MVFormer-T, and 0.3 and 0.4 for MVFormer-S and MVFormer-B, respectively.
We fine-tune the models trained at 224$\times$224 resolution for 30 epochs using exponential moving average~\cite{polyak1992acceleration} for 384$\times$384 resolution.
The proposed implementation is based on PyTorch library~\cite{paszke2019pytorch}, and the experiments are run on 8 NVIDIA A100 GPUs.

Table~\ref{tab:sota_in1k} presents performance comparisons of MVFormer with the current SOTA models on ImageNet-1K classification. 
We compare MVFormer to existing attention-based~\cite{liu2021swin,touvron2021training,yang2021focal,wang2021pyramid} and convolution-based~\cite{liu2022convnet,yu2023inceptionnext,yang2022focal,guo2022visual,yu2022metaformer1} 
% and MLP-based~\cite{wei2022activemlp,chen2021cyclemlp,touvron2021resmlp,zhang2022morphmlp}
SOTA models, grouped into the model size represented by the number of parameters and MACs.
Throughout both approaches, the MVFormer variants consistently outperform other candidates. Particularly, MVFormer-T, S, and B beat the current convolution-based SOTA models, ConvFormer-S18, S36, and M36, by 0.4\%p, 0.2\%p and 0.1\%p, respectively, regarding performance enhancements with equal or fewer parameters and MACs.
On higher-resolution images, the performance increases occur in all three model variants.
% Especially, MVFormer-T, -S and -B beat current convolution-based SOTA models, ConvFormer-S18, -S36 and -M36, with 0.4\%p, 0.2\%p and 0.1\%p of performance enhancements with equal or fewer parameters and MACs. 

% Particularly, MVFormer-S exhibits powerful efficiency that it even attains competitive performance compared to other -B models, as saving approximately half of parameters and FLOPs (Figure~\ref{fig:plot}). 

\subsection{Object Detection and Instance Segmentation}
\begin{table}[hbt!]
    \centering\resizebox{\linewidth}{!}{\setlength\tabcolsep{0pt}

    \begin{tabular}{l|cc|ccc|ccc}
        \Xhline{1.0pt}
    	\multirow{2}{*}{Method}& $\#$Params & MACs  &   \multirow{2}{*}{$\text{AP}^\text{b}$} &\multirow{2}{*}{$\text{AP}^\text{b}_{50}$}&\multirow{2}{*}{$\text{AP}^\text{b}_{75}$}&\multirow{2}{*}{$\text{AP}^\text{m}$}&\multirow{2}{*}{$\text{AP}^\text{m}_{50}$}&\multirow{2}{*}{$\text{AP}^\text{m}_{75}$} \\
    		 & (M) & (G) & &  &  & & \\
           	\Xhline{1.0pt} 
            \multicolumn{9}{c}{Mask R-CNN 1$\times$}\\
                \hline 
                PVT-Small~\cite{wang2021pyramid} & 44 & 245 & 40.4  & 62.9 & 43.8 & 37.8 & 60.1 & 40.3\\ 
                PoolFormer-S36~\cite{yu2022metaformer}  & 51 & 266 & 41.0 & 63.1 & 44.8 & 37.7 & 60.1& 40.0\\ 
                Swin-T~\cite{liu2021swin} & 48 & 264  & 42.2 & 64.6 & 46.2 & 39.1 & 61.6 & 42.0 \\
                 Focal-T~\cite{yang2021focal}  & 49 & 291 & 44.8 & 67.7 & 49.2 & 41.0 & 64.7& 44.2\\ 
                \rowcolor{LightCyan}
                \textbf{MVFormer-T}& \textbf{44} &  \textbf{231} & \textbf{46.2} &   \textbf{68.4} &   \textbf{50.8} &\textbf{42.1} & \textbf{65.4} & \textbf{45.2} \\     
                \hline 
                PVT-Medium~\cite{wang2021pyramid} & 64 & 295 & 42.0 & 64.4 & 45.6 & 39.0 & 61.6  & 42.1 \\ 
                PVT-Large~\cite{wang2021pyramid} & 81 & 364 & 42.9 & 65.0 & 46.6 & 39.5 & 61.9  & 42.5 \\ 
                Swin-S~\cite{liu2021swin} & 69 & 364 & 46.5 & 68.7 & 51.3 & 42.1 & 65.8 & 45.2 \\
                 Focal-S~\cite{yang2021focal}  & 71 & 401 & 47.4 & \textbf{69.8} & 51.9 & 42.8 &66.6 & 46.1\\ 
                \rowcolor{LightCyan}
    		 \textbf{MVFormer-S}&   \textbf{57} & \textbf{299}   &  \textbf{47.5}  &\textbf{ 69.8}& \textbf{52.3} &\textbf{43.0} & \textbf{66.8}&\textbf{46.3} \\     
            \hline
                \multicolumn{9}{c}{RetinaNet 1$\times$}\\
                \hline  
                PoolFormer-S36~\cite{yu2022metaformer}  & 41 & - & 39.5 & 60.5 & 41.8 & - & -&-\\ 
                PVT-Small~\cite{wang2021pyramid}  & 34 & - & 40.4 & 61.3 & 43.0 & - &-&-\\ 
                 Focal-T~\cite{yang2021focal}  & 39 & 265 & 43.7 & - & - & - & -& -\\ 
            \rowcolor{LightCyan}
    		 \textbf{MVFormer-T}& \textbf{34} &  \textbf{231} & \textbf{44.8} &   \textbf{66.1} &   \textbf{48.0} &- & - & - \\   
                \hline 
                PVT-Medium~\cite{wang2021pyramid}  & 54 & - & 41.9 & 63.1 & 44.3 & - & -&-\\ 
                PVT-Large~\cite{wang2021pyramid}  & 71 & - & 42.6 & 63.7 & 45.4& - & -&-\\ 
                 Focal-S~\cite{yang2021focal}  & 62 & 367 & 45.6 & - & - & - & -& -\\ 
                \rowcolor{LightCyan}
    		 \textbf{MVFormer-S}&   \textbf{47} & \textbf{306}   & \textbf{45.8}  & \textbf{67.2}& \textbf{49.3} & -& -& -\\   
            \hline
                \multicolumn{9}{c}{Mask R-CNN 3$\times$}\\
                \hline
            Swin-T~\cite{liu2021swin}  & 48 & 264 & 46.0 & 68.1 & 50.3 & 41.6 & 65.1 & 44.9\\ 
            ConvNeXt-T~\cite{liu2022convnet}  & 48 & 262 & 46.2 & 67.9 & 50.8 & 41.7 & 65.0 & 44.9\\ 
                 Focal-T~\cite{yang2021focal}  & 49 & 291 & 47.2 & \textbf{69.4} & 51.9 & 42.7 & \textbf{66.5}& 45.9\\ 
                \rowcolor{LightCyan}
    		 \textbf{MVFormer-T}& \textbf{44} &  \textbf{231} & \textbf{47.6}  & 69.1& \textbf{52.0} & \textbf{43.1}&66.2& \textbf{46.5}\\   
            \Xhline{1.0pt}
    \end{tabular}
    }
\caption{\textbf{Object detection and instance segmentation performance on COCO val2017.} MACs are calculated with the input shape of 1280$\times$800.}
\label{tab:sota_coco}
% \vspace*{-.2cm}
    
\end{table}  
\noindent We evaluate MVFormer regarding object detection and instance segmentation tasks on the COCO 2017 benchmark~\cite{lin2014microsoft}, with 118K training images and 5K validation images.
We use the ImageNet-1K pre-trained MVFormer as the backbone, which is equipped with the Mask R-CNN~\cite{he2018mask} and RetinaNet~\cite{lin2018focal}.
% Cascade Mask R-CNN~\cite{cai2019cascade},
% For a fair comparison, we followed the training strategy of Swin~\cite{liu2021swin} for both 1$\times$ and 3$\times$ schedule.
We train the model with single-scale inputs with a learning rate of 1e-4 for RetinaNet and 2e-4 for Mask R-CNN, decayed at 8 and 11 for 1$\times$ schedule, and at 27 and 33 for 3$\times$ schedule. 
The image is resized to the shorter side, at 800 pixels, whereas the longer side remains within the limit of 1333 pixels. 
% The shorter side of the images is also resized to 800 pixels for testing. 
To prevent overfitting, MVFormer-T and S have stochastic depths set to 0.3 and 0.4, respectively.
% Experiments are performed using 8 NVIDIA A100 GPUs, and 
The implementation is based on the mmdetection~\cite{chen2019mmdetection}.

Table~\ref{tab:sota_coco} presents the performance comparison of MVFormer with SOTA ViT models.
% MLP-based~\cite{chen2021cyclemlp,wei2022activemlp}, attention-based~\cite{liu2021swin,wang2021pyramid} and convolution-based~\cite{chen2023run,yang2022focal} SOTA ViT models.
%Both MVFormer-T and MVFormer-S show impressive efficiencies, as individually achieving competitive performances with significantly less parameters and FLOPs.
In all cases, our MVFormer-T and MVFormer-S consistently achieve SOTA performances with the highest mean average precision (mAP) on both tasks, with significantly fewer parameters and MACs.
For 1$\times$ schedule, MVFormer-variants even present best $\mathrm{mAP_{50}}$ and $\mathrm{mAP_{75}}$ with both Mask R-CNN and RetinaNet.
% This result suggests the great generalization of MVFormer.
This result underscores the exceptional generalization capability of MVFormer.
In the case of 3$\times$ schedule, compared to Focal-T~\cite{yang2021focal}, MVFormer-T shows slightly lower $\mathrm{mAP_{50}}$ on both tasks. 
Nevertheless, considering higher mAP and $\mathrm{mAP_{75}}$, it becomes evident that MVFormer excels in providing more precise dense predictions.
% it is shown that the MVFormer performs more precise dense prediction.

\subsection{Semantic Segmentation}
\begin{table}[hbt!]
	\centering\renewcommand{\arraystretch}{1.}\resizebox{\linewidth}{!}{
    \setlength\tabcolsep{6pt}

    \begin{tabular}{l|c|c|c}
        \Xhline{1.0pt}
        \multirow{2}{*}{Method} & \multicolumn{3}{c}{Semantic FPN} \\
        \cline{2-4}
         &\#Params(M) & MACs(G) & mIoU(\%) \\
                \Xhline{1.0pt}
        % {Method}
        %  &\#Params(M) & MACs(G) & mIoU(\%) \\
        
        %     \Xhline{1.0pt}
        %     \multicolumn{4}{c}{Semantic FPN}\\
        %     \hline
        % ResNet-50& 29  & 46 & 36.7 \\
        PVT-Small~\cite{wang2021pyramid}& 28  & 45 & 39.8 \\
        PoolFormer-S36~\cite{yu2022metaformer}  & 35 & 48  & 42.0 \\
        % Swin-T& - & -  & - \\
        % CycleMLP-B2~\cite{chen2021cyclemlp} & 31 & 42 & 43.4 \\
        % ATMNet-T~\cite{wei2022activemlp} & 31 & 42 & 45.8 \\
        InceptionNeXt-T~\cite{yu2023inceptionnext}& 28  & 44 & 43.1 \\
        VAN-B2~\cite{guo2022visual} &30&48&46.7\\
        % \hline
        \rowcolor{LightCyan}
        \textbf{MVFormer-T} & \textbf{28}  & \textbf{42} &\textbf{47.1} \\
        \hline
        % ResNet-101& 48  & 65 & 38.8 \\
        % PVT-Medium ~\cite{wang2021pyramid} & 48  & 61 & 41.6 \\
        PoolFormer-M36~\cite{yu2022metaformer}  & 60 & 68  & 42.4 \\
        PVT-Large ~\cite{wang2021pyramid} & 65  & 80 & 44.8 \\
        % Swin-S& - & -  & - \\
        % CycleMLP-B3~\cite{chen2021cyclemlp} & 42 & 58 & 44.3 \\

        % ATMNet-B~\cite{wei2022activemlp}& 56 & 75 & 47.7 \\
        InceptionNeXt-S~\cite{yu2023inceptionnext}& 50  & 65 & 45.6 \\
        VAN-B3~\cite{guo2022visual}&49&68&48.1\\
        % \hline
        \rowcolor{LightCyan}
        \textbf{MVFormer-S} &  \textbf{41} & \textbf{61} & \textbf{48.8} \\            \hline
            % \multicolumn{4}{c}{UperNet} \\
            % \hline
        %     Swin-T~\cite{liu2021swin}   &  60  & 945 & 45.8 \\ 
        %         ConvNeXt-T~\cite{liu2022convnet}  & 60  & 939 & 46.7 \\
        %      InceptionNeXt-T~\cite{yu2023inceptionnext}& 56  & 933 & 47.9 \\
        %      VAN-B2~\cite{guo2022visual}&57&948&50.1\\
        %                  \rowcolor{LightCyan}
        % \textbf{MVFormer-T} & \textbf{53}  & \textbf{925} &\textbf{} \\
        % \hline
        
        %     Swin-S~\cite{liu2021swin}   &  81  & 1038 & 49.5 \\ 
        %         ConvNeXt-S~\cite{liu2022convnet}  & 82  & 1027 & 49.6 \\
        %      InceptionNeXt-S~\cite{yu2023inceptionnext}& 78  & 1020 & 50.6 \\
        %      VAN-B3~\cite{guo2022visual}&75&1030&50.6\\
        %     \rowcolor{LightCyan}
        % \textbf{MVFormer-S} &  \textbf{66} & \textbf{1002} & \textbf{} \\ 
        \Xhline{1.0pt}
    \end{tabular}}
    
    \caption{\textbf{Semantic segmentation performance on ADE20K validation set.} MACs are calculated with the input shape of 512$\times$512.}
    \label{tab:semantic_segmentation}
% \vspace*{-.2cm}
\end{table}

We also assess MVFormer on semantic segmentation using ADE20K benchmark~\cite{zhou2019semantic},
comprising 20K training and 2K validation images. 
We employ the ImageNet-1K pre-trained MVFormer as the backbone, equipped with the Semantic FPN~\cite{kirillov2019panoptic}.
% and UperNet~\cite{xiao2018unified}. 
To train 40K iterations with a batch size of 32, we use AdamW with an initial learning rate of $3\times10^{-4}$ and cosine learning rate schedule. Images are resized and cropped to $512\times512$ pixels for training.
% and 512 pixels on the shorter side for testing. 
% Experiments are performed using 8 NVIDIA A100 GPUs, and 
The implementation is based on mmsegmentation~\cite{contributors2020mmsegmentation}.

In Table~\ref{tab:semantic_segmentation}, we compare MVFormer with SOTA models for the semantic segmentation task.
%we present the performance comparison of MVFormer with SOTA models for the semantic segmentation task.
%Table~\ref{tab:semantic_segmentation} shows the performance comparison of MVFormer with SOTA models.
% Similar to the results in object detection and instance segmentation,
Both MVFormer-T and MVFormer-S highly outperform other models given a competitive number of parameters and MACs.
Compared to the VAN-B2 and B3~\cite{guo2022visual}, which are up-to-date convolution-based ViTs, MVFormer-T and S display 0.4\%p and 0.7\%p performance gains, respectively, with better efficiency.

\subsection{Ablation Studies} 
\noindent We perform ablation studies to validate the effectiveness of MVN and MVTM.
All experiments are conducted on the ImageNet-1K classification, with the MVFormer-xT model.
% This section aims to examine the factors that contribute to the good performance of MVFormer. We first analyze the Tri-Norm and MVTM. All experiments were conducted on ImageNet 1K.

\subsubsection{Ablation Study on Individual Modules}
\begin{table}[hbt!]
	\centering\resizebox{\linewidth}{!}{
    \setlength\tabcolsep{8pt}
    \begin{tabular}{cc|c|c|c}
        \Xhline{1.0pt}
 % MVTM & MVN_{N_1} & MVN_{N_2} & \#Params & MACs  & Top-1 Acc.$(\%)$ \\
  MVTM & MVN & \#Params & MACs  & Top-1 Acc.$(\%)$ \\

\hline
 &    & 16.88M & 2.17G & 80.63 \\
 % & $\checkmark$&  &  16.89M & 2.18G & 80.91 (+0.23) \\
  % & & $\checkmark$ & 16.89M &2.18G & 81.12 (+0.44) \\
  $\checkmark$ &  &  16.94M & 2.19G & 80.80 (+0.17) \\
   & $\checkmark$ & 16.90M & 2.18G & 81.16 (+0.53) \\
 % $\checkmark$ & $\checkmark$ & &16.95M & 2.19G & 80.85 (+0.17) \\
 %   $\checkmark$ & & $\checkmark$ & 16.95M &2.19G & 81.05 (+0.37) \\
               \hline
$\checkmark$   & $\checkmark$ 
 & 16.96M & 2.19G & \textbf{81.30 (+0.67)} \\
        \Xhline{1.0pt}
    \end{tabular}}
%    \caption{Ablation study of MVN and MVTM. }
    \caption{\textbf{Ablation results of the proposed modules, MVN and MVTM.} }
    % $\text{MVN}_{N_1}$ adopts MVN in the front token mixer, while $\text{MVN}_{N_2}$ employs MVN in the front channel mixer.}
    \label{tab:ab_main}
% \vspace*{-.2cm}
\end{table}  

%We conduct basic ablation experiment to evaluate the effectiveness of our MVN and MVTM.
\noindent We conduct ablation experiments in Table~\ref{tab:ab_main} to evaluate the effect of each proposed module with a convolution-based ViT baseline on the ImageNet-1K classification.
For a fair comparison, we design a MetaFormer-based baseline with a token mixer equal to the 5$\times$5 depthwise separable convolution 
% since the MVTM consists of multi-scale depthwise convolution followed by a pointwise convolution, and
because it requires a similar number of parameters and MACs compared to the MVTM. 
Regarding normalization, LN is applied by default.
% As listed in Table~\ref{tab:ab_main}, $\text{MVN}_{N_1}$ ($\text{MVN}_{N_2}$) denotes the use of MVN in the first (second) normalization position within the MetaFormer block.
% MVN was partitioned into two components, $\text{MVN}_{N_1}$ and $\text{MVN}_{N_2}$. 
% Individually applying $\text{MVN}_{N_1}$ and $\text{MVN}_{N_2}$, produced 0.23\%p and 44\%p improvements in performance, respectively, increasing to 0.5\%p when both are applied.
% However, applying both $\text{MVN}_{N_1}$ and $\text{MVN}_{N_2}$ resulted in a performance improvement of 0.5\%p. 
% Employing either MVN or MVTM individually results in performance enhancements of over 0.12\%p and 0.50\%p, respectively, with a negligible increase in the number of parameters and MACs.
% For MVTM, performance is boosted at 0.12\%p, and with $\text{MVN}_{N_1}$ and $\text{MVN}_{N_2}$, the improvement increases to 0.17\%p and 0.34\%p, respectively.
% Additionally, employing either MVN or MVTM individually results in performance enhancements of over 0.5\%p and 0.12\%p, respectively, with a negligible increase in the number of parameters and MACs.
When each of the MVN and MVTM is solely used, there occur significant performance enhancements of 0.53\%p and 0.17\%p with negligible amount of additional parameters and MACs, respectively.
Between them, the MVN improves more performance of 0.38\%p than MVTM.
% @@ 윗 문장: As shown in Table~\ref{tab:ab_basic}, employing either MVN or MVTM individually results in marginal performance enhancements of over 0.8% and 0.5%, respectively, with a negligible increase in the number of parameters and FLOPs
Additionally, MVFormer-xT, which incorporates MVN and MVTM, achieves the highest performance of 81.30\%. 
%These results clearly support the module-level benefits of the MVN and MVTM on model performance.
These findings support the combined use of these proposed modules and the individual benefits each module has in improving model performance.
% These results support not only the combined use of these proposed modules but also the distinct advantages of each individual module in enhancing model performance.
% \input{tables/ablation_basic} 

\subsubsection{Various Combinations of Three Normalizations}
\begin{table}[hbt!]
	\centering
    \setlength\tabcolsep{5pt}\renewcommand{\arraystretch}{1.}\resizebox{\linewidth}{!}{

    \begin{tabular}{c|c|c|c|c}
        \Xhline{1.0pt}
        Method & Type &\#Params&MACs & Top-1 acc.\\
        % \hline
        % None & - &14.84M&1.87G& 78.46\%\\
        \hline
        LN (Baseline) &  Uni & 16.95M& 2.19G& 80.75\%\\
        BN & Uni & 16.95M & 2.19G & 80.67\%\\
        IN & Uni & 16.95M & 2.19G & 78.83\%\\
        \hline
        LN + BN & Bi &16.96M&2.19G&80.86\%\\
        LN + IN & Bi &16.96M&2.19G&
        81.17\%\\
        BN + IN & Bi &16.96M&2.19G&81.01\%\\
        \hline
        % LN + BN + IN (Avg.) & Tri &14.88M&1.87G&\%\\
        LN + BN + IN (Ours) & Tri &16.96M&2.19G&\textbf{81.30\%}\\
        \Xhline{1.0pt}
    \end{tabular}}
    \caption{\textbf{Ablation study of normalization methods in MVN.}}
    \label{tab:ab_norm}
% \vspace*{-.2cm}
\end{table}  

\noindent Table~\ref{tab:ab_norm} presents the ablation study of all combinations of the three normalization methods in MVN.
Combining just two normalized features consistently enhances the performance compared to that of a single method.
In particular, IN significantly degrades the performance when used alone.
%, even lower than when compared to using LN. 
Nevertheless, IN exhibits beneficial synergy when combined with other methods.
We infer that IN contributes to the performance gains by mitigating the batch-dependence in BN and spatial distribution variation in LN.
%Our MVN, in which BN, LN and IN are combined, marginally outperforms all other combinations,
%which  strongly supports our expectation that comprehensively encompassing diverse characteristics of normalization methods leads to improved performance, contributing to an enhanced perspective on extending feature diversity.
The MVN, combining BN, LN and IN, significantly outperforms all other combinations, strongly supporting the conjecture that comprehensively encompassing diverse characteristics of normalization methods leads to improved performance and contributes to an enhanced perspective on extending feature diversity.
% We evaluated the performance of applying LN, BN and IN across channels in order to confirm the effectiveness of each normalization technique and we also examined how each norm affects performance individually. To analyze the impact of the TriNorm components, we evaluated how the performance of the model changed as Uni-, Bi- or Tri- types LN, BN, and IN were applied. Note that the adoption of the tri-normalization technique leads to increased feature diversity. (See Table~\ref{tab:ab_norm})

\subsubsection{MVN on Existing ViT and CNN Models} 
\begin{table}[hbt!]
	\centering
    \setlength\tabcolsep{4pt}\renewcommand{\arraystretch}{.8}\resizebox{\linewidth}{!}{

    \begin{tabular}{c|c|c|c|c}
        \Xhline{1.0pt}
        Model & Method & \#Params & MACs & Top-1 acc. \\
        \hline
        \multirow{2}{*}{Swin-T~\cite{liu2021swin}} & LN & 28M & 4.4G  & 81.2\%\\
        %Poolformer-s12 & Bi(LN+BN) & 12M & 1.83G & \textbf{77.5\%} \\ 
        & MVN & 28M & 4.4G  &  \textbf{81.4\%(+0.2)} \\
        \hline
        \multirow{2}{*}{ConvFormer-S18~\cite{yu2022metaformer1}} & LN & 27M & 3.9G  & 83.0\%\\
        %Poolformer-s12 & Bi(LN+BN) & 12M & 1.83G & \textbf{77.5\%} \\ ConvFormer-S18
        & MVN & 27M & 3.9G  &  \textbf{83.2\%(+0.2)} \\
        \hline
        \multirow{2}{*}{ConvNeXt-T~\cite{liu2022convnet}} & LN &  28M & 4.5G  & 82.1\% \\   
         & MVN &  29M & 4.5G  & \textbf{82.3\%(+0.2)} \\
        
        \hline
        
        \multirow{2}{*}{Poolformer-S36~\cite{yu2022metaformer}} & LN & 31M & 5.0G  & 81.4\% \\
        %Poolformer-s12 & Bi(LN+BN) & 12M & 1.83G & \textbf{77.5\%} \\
         & MVN & 31M & 5.0G & \textbf{81.6\%(+0.2)} \\
        \hline

        \multirow{2}{*}{ResNet50~\cite{he2016deep}} & BN & 26M & 4.1G  & 76.1\%\\ %76.1
        %Poolformer-s12 & Bi(LN+BN) & 12M & 1.83G & \textbf{77.5\%} \\

        %Poolformer-s12 & Bi(LN+BN) & 12M & 1.83G & \textbf{77.5\%} \\
        & MVN & 26M & 4.1G  &  \textbf{76.3\%(+0.2)} \\
        \hline

        \Xhline{1.0pt}
    \end{tabular}}
    
    \caption{
    \textbf{MVN generalization on ViT and CNN models.}
    % Generalization performances of 
    % MVN on existing ViT and CNN models.
    } 
    \label{tab:ab_poolformer}
%\vspace*{-.2cm}
\end{table}  

\noindent To evaluate the generalization of MVN, we apply MVN to existing variants of ViT and CNN.
For ViT candidates, we select Swin~\cite{liu2021swin}, ConvFormer~\cite{yu2022metaformer1}, ConvNeXt~\cite{liu2022convnet}, and PoolFormer~\cite{yu2022metaformer}, which are attention-, convolution- and pooling-based models, respectively, and we experiment on ResNet~\cite{he2016deep}, which represents the CNN.
For ViTs, we substitute the LN with MVN within each block, and in ResNet, all BN layers are replaced with MVNs. 
% For a fair comparison, all presented performances are based on our own implementations.
As listed in Table~\ref{tab:ab_poolformer}, MVN displays impressive generalization, significantly improving the original performance of all five models.
For ViTs, the model achieves 0.2\%p of consistent Top-1 accuracy gains on PoolFormer-S36, Swin-T, ConvFormer-S18 models and ConvNeXt-T. 
%On ConvNeXt-T, the model performance increases by 0.3\%p with almost the same parameters and MACs.
In the case of CNN, MVN  even works on ResNet50 with 0.2\%p of accuracy improvement. 
These results suggest that MVN is not restricted to just CNN-ViT hybrid architecture, indicating promising feasibility for applications in various standard vision models. 
% Regarding the result of ConvFormer-S18 based on MetaFormer-based baseline, we anticipate that its modest performance enhancement (0.2\%p) is influenced by MVN. Nevertheless, through its integration with MVTM, MVFormer-T achieved further enhancements (0.4\%p) in performance (Table~\ref{tab:sota_in1k}).

% Regarding the result of ConvNeXt, we expect that its slight gain of performance affected by the relatively smaller number of normalization layers, as it contains just single normalization layer in a block. Compared to two in other ViTs and three in ResNet, it seems that MVN less contributes to the model performance because of its tiny portion in overall network.

% Although it is relatively smaller than in ViTs, these results support the extensibility of MVN to wide range of common vision architectures.

% We can effectively substitute LN with TriNorm in PoolFormer and improve its performance on ImageNet-1K. Table~\ref{tab:ab_poolformer} shows that PoolFormer baseline can benefit significantly from TriNorm with an accuracy increase of 1.1\%.

\subsubsection{Ablation Experiment on MVTM}
\begin{table}[hbt!]
	\centering
    \setlength\tabcolsep{3pt}\renewcommand{\arraystretch}{1.}\resizebox{\linewidth}{!}{
    \begin{tabular}{l|c|c|c}
        \Xhline{1.0pt}
        Type & \#Params & MACs & Top-1 acc. \\
        
        \Xhline{1.0pt}
        % Single-scale  & 14.86M & 1.87G & 79.75\% \\
        % + Multi-scale  & 14.88M & 1.87G & 79.81\% \\
        % + Stage-spec. channel ratio& 14.88M & 1.87G &  \\
        MVFormer-xT & 16.96M & 2.19G & \textbf{81.30\%} \\
        \hline
       \multicolumn{1}{l|}{(-) Stage-spec. channel split ratio} & 16.94M & 2.20G & 81.20\% \\
       \multicolumn{1}{l|}{(-) Stage-spec. global filter size} & 16.93M & 2.19G & 81.22\% \\
        (-) Stage-spec. both  & 16.92M & 2.19G & 81.15\% \\
       %  \hline
       %  \multicolumn{1}{l|}{(-) Local filter} & 14.92M & 1.89G & 79.88\% \\
       % \multicolumn{1}{l|}{(-) Intermediate filter} & 14.86M & 1.86G & 79.73\% \\
       % \multicolumn{1}{l|}{(-) Global filter} & 14.91M & 1.88G & 79.85\% \\
       % Single scale $5\times5$& 14.86M & 1.87G & 79.75\% \\
        \hline
        \multicolumn{1}{l|}{(-) Local filter} & 16.97M & 2.21G & 80.99\% \\
       \multicolumn{1}{l|}{(-) Intermediate filter} &16.91M & 2.16G & 81.13\% \\
       \multicolumn{1}{l|}{(-) Global filter} & 16.96M &2.19G & 81.22\% \\
        \Xhline{1.0pt}
    \end{tabular}
    }
    \caption{\textbf{Ablation study of MVTM.} 
    When excluding the adjustment of the channel split ratio, that of Stages 2 and 3 is applied at all stages, and when removing the adjustment of the global filter size, that of Stage 3 is equally used. 
    When eliminating the intermediate filter, the ratios of the local and global filters are doubled. When eliminating either the local or global filters, the missing parts are complemented by the intermediate filter.} 
    \label{tab:ab_convnext}
% \vspace*{-.2cm}
\end{table}  
\noindent Table~\ref{tab:ab_convnext} presents the performance variation, excluding each proposed component of the MVTM individually.
% On average, the elimination of stage-specificity much affect performance decrease than that of mixing filters.
In terms of stage specificity, given either a single global filter size or fixed channelwise split ratio, there occurs similar degree of performance degradations, which is 0.10\%p and 0.08\%p, respectively.
When both are applied, it gets larger that 0.15\%p of performance drop is observed.
This result shows that adopting stage specificity enhances the efficiency of MVFormer, as improving the performance given similar MACs and parameters.
% In terms of the stage-specificity, given either of single global filter size or fixed channel-wise split ratio, even both, there occurs similar degree of performance degradations, which is about 0.18\%p.
% This result tells us that both components are effective when used together, and each hardly reflect the stage-specificity by itself.
% As mention in Method section, we expect that specification of appropriate global mixing range at each stage is important for better performance, given preferred 
In addition, the ablation result on the mixing filter presents the importance of different mixing filter levels.
When one of the three mixing filters is excluded, performance degradation occurs consistently.
It is much higher when the smaller size of filter is eliminated.
% , compared to intermediate or global mixing filters.
% Especially, it is much more when the local filter is eliminated, compared to intermediate or global mixing filters.
We infer that this is because the repeated small filters are able to cover a wide range of visual patterns, whereas the large filter struggles to focus on the local area.

\subsubsection{Learned Weights of MVN}
\begin{figure}[ht]
  \centering
  \includegraphics[width=\linewidth]{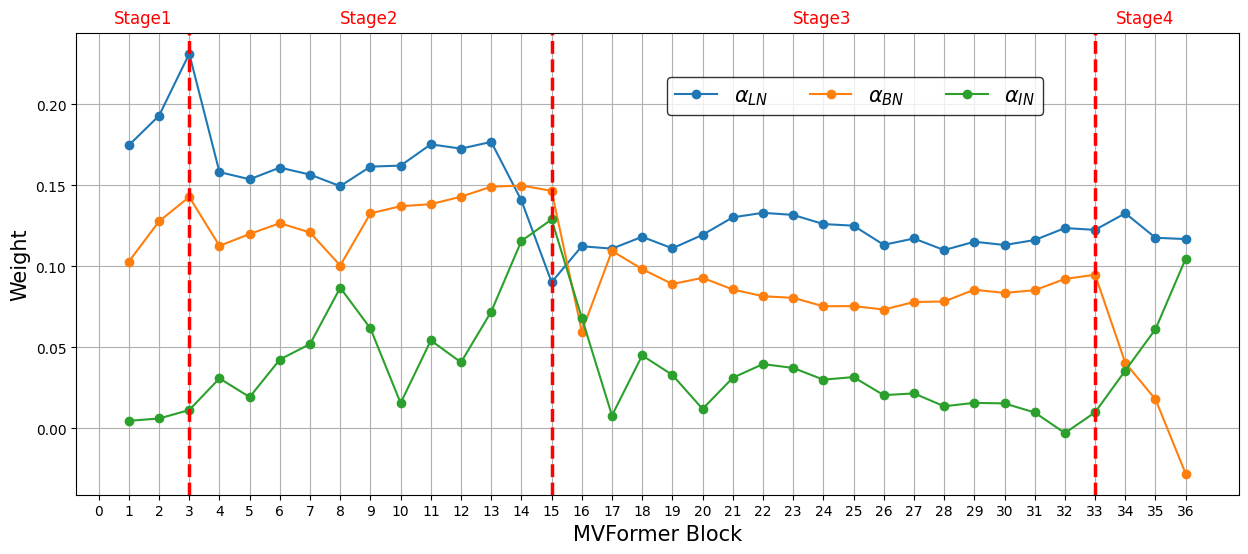}
     % {figures/chart_analysis.pdf}
    
    \caption{\textbf{The average values of $\alpha_{LN}, \alpha_{BN}$ and $\alpha_{IN}$ for each block in MVFormer-S.}} 
    % Across four stages, MVFormer-S consists of 3, 12, 18, and 3 blocks, respectively.}
  \label{fig:analysis_chart}
  % \vspace{-.2cm}
\end{figure}
 % of the Mixing Weights of MVN
% \begin{figure}[ht]
%   \centering
%   \includegraphics[width=\linewidth]{figures/norm_images.png}
%   \begin{tabular}{m{19mm}m{18mm}m{18mm}m{18mm}}
%        (a) Input&(b) BN&(c) LN&(d) IN 
%   \end{tabular}
%   \caption{%
%   Visualization of normalized input images shows distinct visual characteristics following BN, LN, and IN. 
%   }
%   \label{fig:norm_images}
% \end{figure}
% \input{tables/weight_analysis}
% We visualized the results of applying BN, LN, and IN on the input image as shown in Figure~\ref{fig:norm_images}.
\noindent Fig.~\ref{fig:analysis_chart} presents the weight distributions in MVN to identify certain preferences depending on the stage number. %the location of MVN in a block, 
% and the receptive field of the following token mixer 
%(Fig.~\ref{fig:analysis_chart}).
Interestingly, it is observed that there exists an overall tendency of the ratio between three normalization methods.
% note that normalization plays a different role at each level.
Across all stages, excluding the last part of the second stage, the weight of the LN consistently has the highest ratio. 
% However, it can be observed that as the stage increases, the ratio of LN decreases. 
This suggests that the model predominantly reflects the input channel distribution of each pixel, rather than the spatial distribution of each channel. The BN and IN temporarily exhibit a higher ratio than the LN in the final block of Stage 2, possibly due to the model prioritizing spatial over channel distribution during a rapid channel dimension change. Moreover, IN generally has a lower ratio than BN, except in the last stage, suggesting a preference for batch-independent sample-level spatial information.
% Between BN and IN, IN overally exhibits a smaller ratio than BN, except in the last stage. 
% We infer that it is because the classifier prefers batch-independent sample-level spatial information.
% Meanwhile, an distinctive observation is that BN and IN temporarily exhibit a higher ratio than LN in the final block of stage 2.As we conject, considering that there occurs rapid channel dimension leaf of 128 to 320 between stage 2 and 3, the model focus to maintain the spatial distribution of feature rather than the pre-stage channel distribution. 
This observation is consistent in the MVFormer-T and -B models. 

\section{Conclusion}
%In this work, we introduce MVFormer, which effectively learns diverse features by simply combining existing normalization methods and concept
%%%In this work, we propose MVFormer, an efficient yet effective convolution-based ViT, 
%%%to enhance the learning capability of diverse features by introducing a normalization module (MVN) and a token mixer (MVTM) that embrace various perspectives.
This work proposes MVFormer, an efficient yet effective convolution-based ViT, 
by introducing a normalization module (MVN) and a token mixer (MVTM) to extract diverse features from multiple viewpoints.
The MVFormer outperforms the existing SOTA convolution-based ViTs in image classification and three downstream tasks, given competitive efficiency.
%##
%enhance the learning capability of diverse features by introducing a normalization module (MVN) and a token mixer (MVTM) that embrace various perspectives. 
%Our small model achieves competitive performance compared to existing much larger Base models. % ###
%Especially, MVN boosts the performance of MVFormer and notably improves both existing ViT and CNN, affirming its scalability.
Significantly, MVN consistently boosts the performance of MVFormer and existing ViTs and CNNs, affirming its scalability.
We also confirm that triscale filters and stage specificity of the MVTM are crucial factors for performance.
In the future, we will explore the interrelationship between the normalization method and various types of token mixers, and we expect that these approaches offer valuable insights into the vision community and can be extended to other domains.
{
    \small
    \bibliographystyle{ieeenat_fullname}
    \bibliography{main}
}
\clearpage
\setcounter{page}{1}
\renewcommand{\thesection}{\Alph{section}}
\setcounter{section}{0}
\maketitlesupplementary
% \setcounter{table}{8}

% \section{Rationale}
% \label{sec:rationale}
% % 
% Having the supplementary compiled together with the main paper means that:
% % 
% \begin{itemize}
% \item The supplementary can back-reference sections of the main paper, for example, we can refer to \cref{sec:intro};
% \item The main paper can forward reference sub-sections within the supplementary explicitly (e.g. referring to a particular experiment); 
% \item When submitted to arXiv, the supplementary will already included at the end of the paper.
% \end{itemize}
% % 
% To split the supplementary pages from the main paper, you can use \href{https://support.apple.com/en-ca/guide/preview/prvw11793/mac#:~:text=Delete%20a%20page%20from%20a,or%20choose%20Edit%20%3E%20Delete).}{Preview (on macOS)}, \href{https://www.adobe.com/acrobat/how-to/delete-pages-from-pdf.html#:~:text=Choose%20%E2%80%9CTools%E2%80%9D%20%3E%20%E2%80%9COrganize,or%20pages%20from%20the%20file.}{Adobe Acrobat} (on all OSs), as well as \href{https://superuser.com/questions/517986/is-it-possible-to-delete-some-pages-of-a-pdf-document}{command line tools}.

\section{Model Configurations}
\label{model_config}
\begin{table}[hbt!]
\centering
\resizebox{8.4cm}{!}{
\renewcommand{\arraystretch}{1.2}
\begin{tabular}{c|c|c|c|cccc}
\Xhline{1.0pt}
\multirow{2}{*}{Stage} & \multirow{2}{*}{$\#$Tokens} & \multirow{2}{*}{Layer} & \multirow{2}{*}{Type} & \multicolumn{4}{c}{MVFormer}                                                    \\ \cline{5-8} 
                  &                   &                   &                   & \multicolumn{1}{l|}{xT} & \multicolumn{1}{l|}{T} & \multicolumn{1}{l|}{S} & B \\ 
                \Xhline{1.0pt}
\multirow{4}{*}{1} & \multirow{4}{*}{$\frac{H}{4}\times\frac{H}{4}$} & \multirow{2}{*}{Patch Embed.} &      Downsample             & \multicolumn{4}{l}{7$\times$7, \text{stride 4, pad 2}} \\ \cline{4-8} 
                  &                   &                   & Embed. dim                  & \multicolumn{3}{c|}{64}                                            &  96\\ \cline{3-8} 
                  &                   & \multirow{2}{*}{\makecell[c]{MVFormer\\Block}} & MLP ratio& \multicolumn{4}{c}{4}                                                                 \\ \cline{4-8} 
                  &                   &                   &            $\#$ Block      & \multicolumn{1}{c|}{2} & \multicolumn{3}{c}{3} \\
    %              \hline
    %    \multicolumn{8}{c}{Batch Normalization} \\
        \hline
\multirow{4}{*}{2} & \multirow{4}{*}{$\frac{H}{8}\times\frac{H}{8}$} & \multirow{2}{*}{Patch Embed.} &      Downsample             & \multicolumn{4}{l}{3$\times$3, \text{stride 2, pad 1}} \\ \cline{4-8} 
                  &                   &                   &Embed. Dim                  & \multicolumn{3}{c|}{128}                                            &  192\\ \cline{3-8} 
                  &                   & \multirow{2}{*}{\makecell[c]{MVFormer\\Block}} & MLP ratio& \multicolumn{4}{c}{4}                \\ \cline{4-8} 
                  &                   &                   &            $\#$ Block      & \multicolumn{1}{c|}{2} & \multicolumn{1}{c|}{3} & \multicolumn{2}{c}{12}\\
     %             \hline
    %    \multicolumn{8}{c}{Batch Normalization + Layer Normalization} \\
        \hline
\multirow{4}{*}{3} & \multirow{4}{*}{$\frac{H}{16}\times\frac{H}{16}$} & \multirow{2}{*}{Patch Embed.} &      Downsample             & \multicolumn{4}{l}{3$\times$3, \text{stride 2, pad 1}} \\ \cline{4-8} 
                  &                   &                   &Embed. Dim                  & \multicolumn{3}{c|}{320}                                            &  384\\ \cline{3-8} 
                  &                   & \multirow{2}{*}{\makecell[c]{MVFormer\\Block}} & MLP ratio& \multicolumn{4}{c}{4}                   \\ \cline{4-8} 
                  &                   &                   &            $\#$ Block      & \multicolumn{1}{c|}{4} & \multicolumn{1}{c|}{9} & \multicolumn{2}{c}{18}\\
      %            \hline
     %   \multicolumn{8}{c}{Batch Normalization + Layer Normalization} \\     
        \hline
\multirow{4}{*}{4} & \multirow{4}{*}{$\frac{H}{32}\times\frac{H}{32}$} & \multirow{2}{*}{Patch Embed.} &      Downsample             & \multicolumn{4}{l}{3$\times$3, \text{stride 2, pad 1}} \\ \cline{4-8} 
                  &                   &                   &Embed. Dim                  & \multicolumn{3}{c|}{512}                                            &  576\\ \cline{3-8} 
                  &                   & \multirow{2}{*}{ \makecell[c]{MVFormer\\Block}} & MLP ratio & \multicolumn{4}{c}{4}                 \\ \cline{4-8} 
                  &                   &                   &            $\#$ Block      & \multicolumn{1}{c|}{2} & \multicolumn{3}{c}{3} \\
                  \hline
    %    \multicolumn{8}{c}{Layer Normalization} \\
 \multicolumn{4}{c|}{Parameters (M)} & \multicolumn{1}{c|}{17} & \multicolumn{1}{c|}{27} & \multicolumn{1}{c|}{40} & \multicolumn{1}{c}{57} \\
                   \hline

 \multicolumn{4}{c|}{MACs (G)} & \multicolumn{1}{c|}{2.2} & \multicolumn{1}{c|}{3.9} & \multicolumn{1}{c|}{7.6} & \multicolumn{1}{c}{12.7} \\                \Xhline{1.0pt}

\end{tabular}
}
\caption{\textbf{Detailed configurations of MVFormer variant models, MVFormer-xT, T, S, and B.} MACs are calculated with the input
shape of 224×224 on ImageNet-1K classification~\cite{5206848}.}
\label{tab:configuration}
\vspace{-.4cm}
\end{table}

\section{Training Configurations}
% We train our models from the scratch on ImageNet-1K dataset. Table \ref{tab:setup} has further training configurations.
\begin{table}[hbt]
	\centering\renewcommand{\arraystretch}{.7}
      \resizebox{8.4cm}{!}{
    \setlength\tabcolsep{15pt}
    \begin{tabular}{l|c|c|c|c}
        \Xhline{1.0pt}
        \multirow{2}{*}{Configuration }& \multicolumn{4}{c}{MVFormer}\\
        \cline{2-5} 
        &xT&T&S&B\\
        \hline
        Epochs&\multicolumn{4}{c}{300}\\
        Batch size&\multicolumn{4}{c}{4096}\\
        Optimizer&\multicolumn{4}{c}{AdamW}\\
        Learning rate&\multicolumn{4}{c}{4e-3}\\
        Learning rate decay&\multicolumn{4}{c}{Cosine}\\
        Warmup epochs&\multicolumn{4}{c}{20}\\
        Weight decay&\multicolumn{4}{c}{0.05}\\
        Rand Augment&\multicolumn{4}{c}{9/0.5}\\
        Cutmix&\multicolumn{4}{c}{1.0}\\
        Mixup&\multicolumn{4}{c}{0.8}\\
        Label smoothing&\multicolumn{4}{c}{0.1}\\
        Random erasing prob&\multicolumn{4}{c}{0.25}\\
        Peak stochastic depth rate&\multicolumn{4}{c}{0.2 / 0.2 / 0.3 / 0.4}\\
        Head dropout rate&\multicolumn{4}{c}{0.0}\\
        EMA decay rate&\multicolumn{4}{c}{None}\\
        % Peak stochastic depth rate&\multicolumn{1}{c}{0.2}&\multicolumn{1}{c}{0.2}&\multicolumn{1}{c}{0.3}&\multicolumn{1}{c}{0.4}\\
        \Xhline{1.0pt}
    \end{tabular}}
    \caption{\textbf{Training configurations of MVFormer for ImageNet-1K classification.}} 
    \label{tab:setup}
\vspace{-.4cm}
\end{table}  

\section{Learning Curve Analysis}
\begin{figure}[ht]
  \centering
  \includegraphics[width=\linewidth]{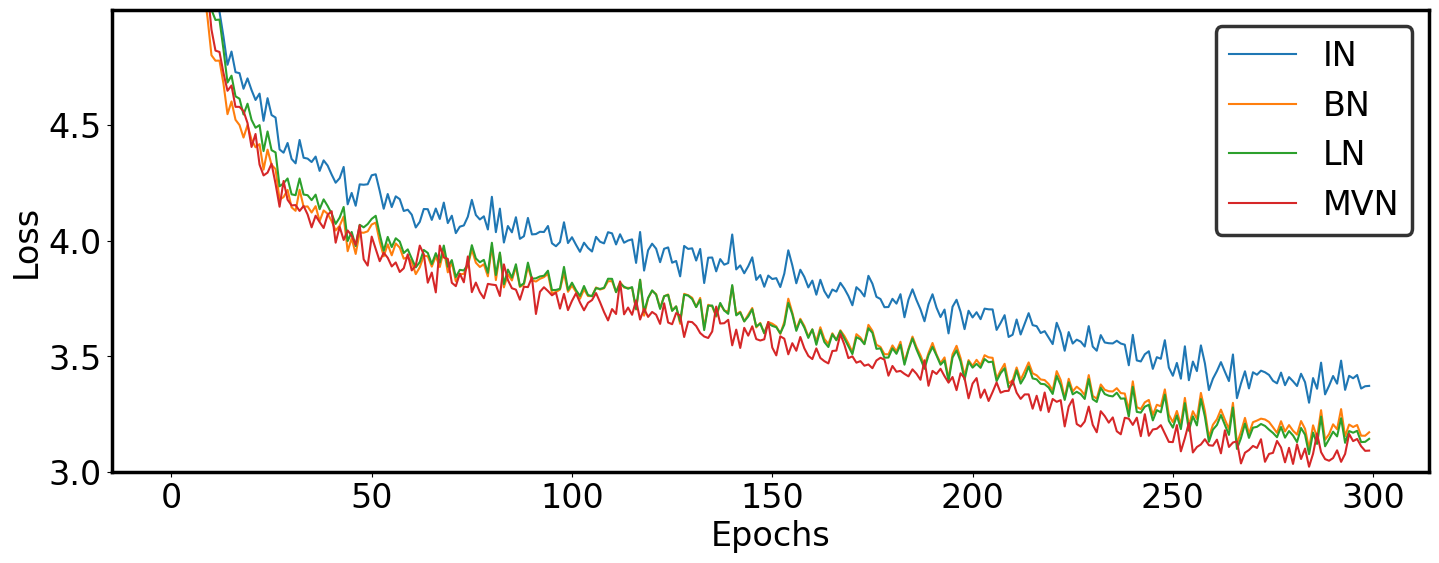}
    \caption{\textbf{Comparison of training loss curves of different normalization methods, BN, LN, IN, and MVN, over 300 epochs.}}
  \label{fig:chart_loss}
  \vspace{-.4cm}
\end{figure}
\noindent We compare the learning curves of MVFormer-xT trained with BN~\cite{ioffe2015batch}, LN~\cite{ba2016layer}, IN~\cite{ulyanov2017instance}, and MVN to identify training stability and convergence rate using each method. 
In Fig.~\ref{fig:chart_loss}, we observe that the overall learning trends of all four cases exhibit similar patterns.
From the perspective of training stability, there are no significant gaps between the four curves, showing similar degrees of oscillation.
In terms of the convergence rate, the training losses of all methods steadily decrease for 300 epochs, where the MVN almost shows the lowest values compared to BN, LN, and IN.
This result suggests the efficacy of MVN that enables the model training to reach to more optimal point while maintaining competitive training stability.
% shows that all the normalization approaches are able to converge, with MVN also conversing in similar pattern.
% and more stable learning. 
% BN and LN mean and variance computation introduces uncertainty caused by the stochastic batch sampling, which helps regularization due to sampling at the layer and batch level. IN computes mean and variance based solely on channels, which can lead to a longer convergence time. Despite incorporating multiple normalization techniques, MVN showed similar learning stability to a single normalization.

\section{Grad-CAM Visualization}
\begin{figure*}[hp!] %ht
  \centering
  \includegraphics[width=15cm,height=22cm,keepaspectratio]{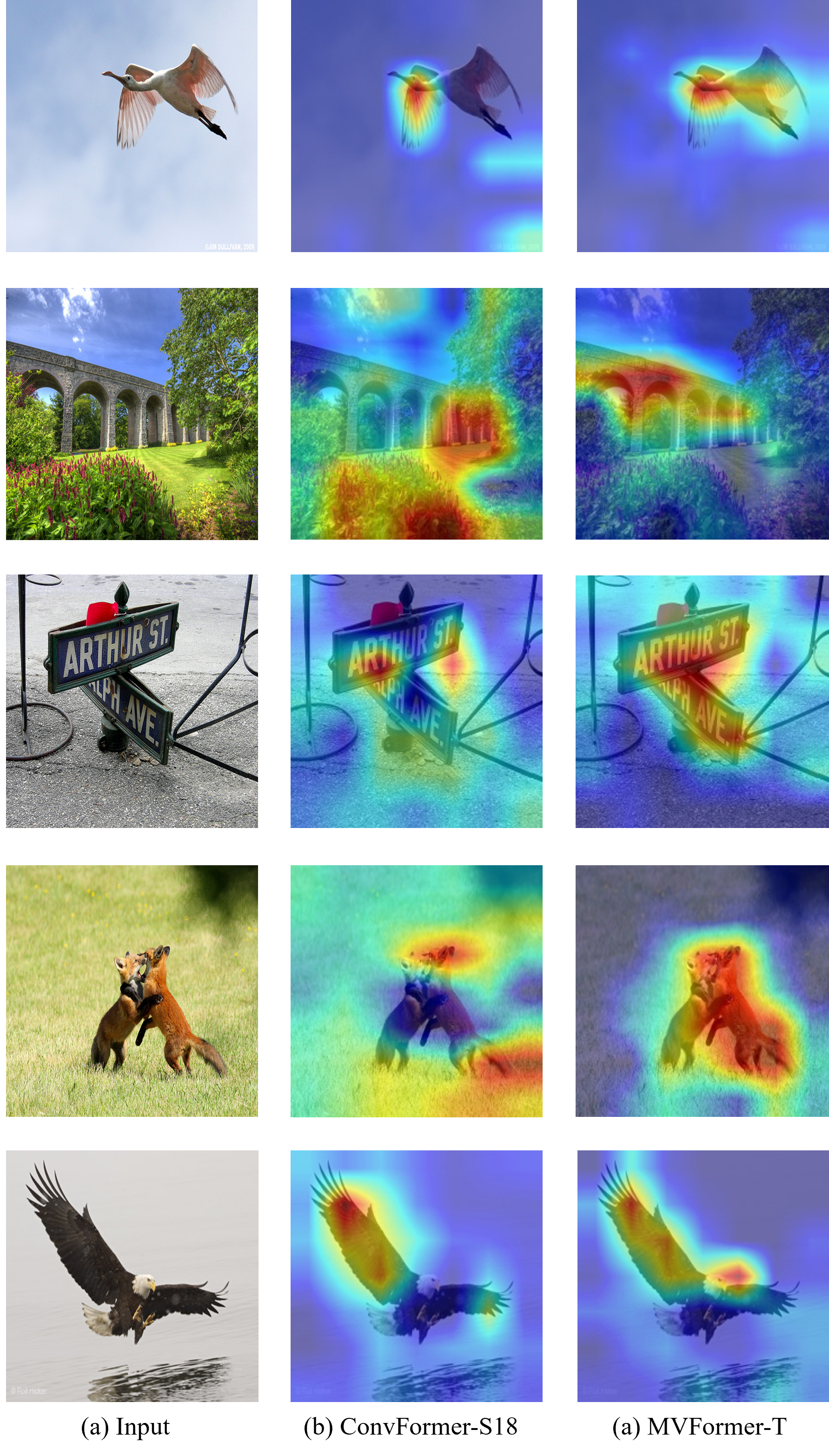}
  % \begin{tabular}{m{19mm}m{18mm}m{14mm}m{20mm}}
  %      (a) Input&(b) BN&(c) LN&(d) BN+LN 
  % \end{tabular}
  \caption{\textbf{Activation maps generated by Grad-CAM for the ConvFormer-S18 and MVFormer-T models.}}
  \label{fig:gradcam}
\end{figure*}

To qualitatively assess the effectiveness of the proposed MVFormer, we conduct a visual comparison with the baseline ConvFormer~\cite{yu2022metaformer1} using Grad-CAM~\cite{selvaraju2017grad} visualization.
% , we use Swin and ConvNeXt, which are currently the most fundamental attention- and convolution-based backbone architectures, respectively.
% We conduct a visual comparison with Swin and ConvNeXt, which are currently the most fundamental attention- and convolution-based ViTs.
Fig.~\ref{fig:gradcam} presents the activation maps of ConvFormer-S18 and MVFormer-T models, both trained on the ImageNet-1K.
% The Grad-CAM~\cite{selvaraju2017grad} activation maps of the Swin-T, ConvNeXt-T, and MVFormer-T models, which are trained on the ImageNet-1K dataset, shown in Figure \ref{fig:gradcam}. 
%Compared to Swin and ConvNeXt, MVFormer is able to locate the objects with greater precision.
Compared to the baseline, our MVFormer effectively captures various scales of main objects in the input images.

% both global and detailed features of the objects.
% the main objects in the input images.

% both global and detailed features of the objects. %enabling localization.
% In the case of the ostrich, MVFormer can see how well it includes even the detailed neck.

\section{Visual Comparison of Normalized Images from BN, LN, IN, and MVN}
\begin{figure*}[hp] %ht
  \centering
  \includegraphics[width=\linewidth]{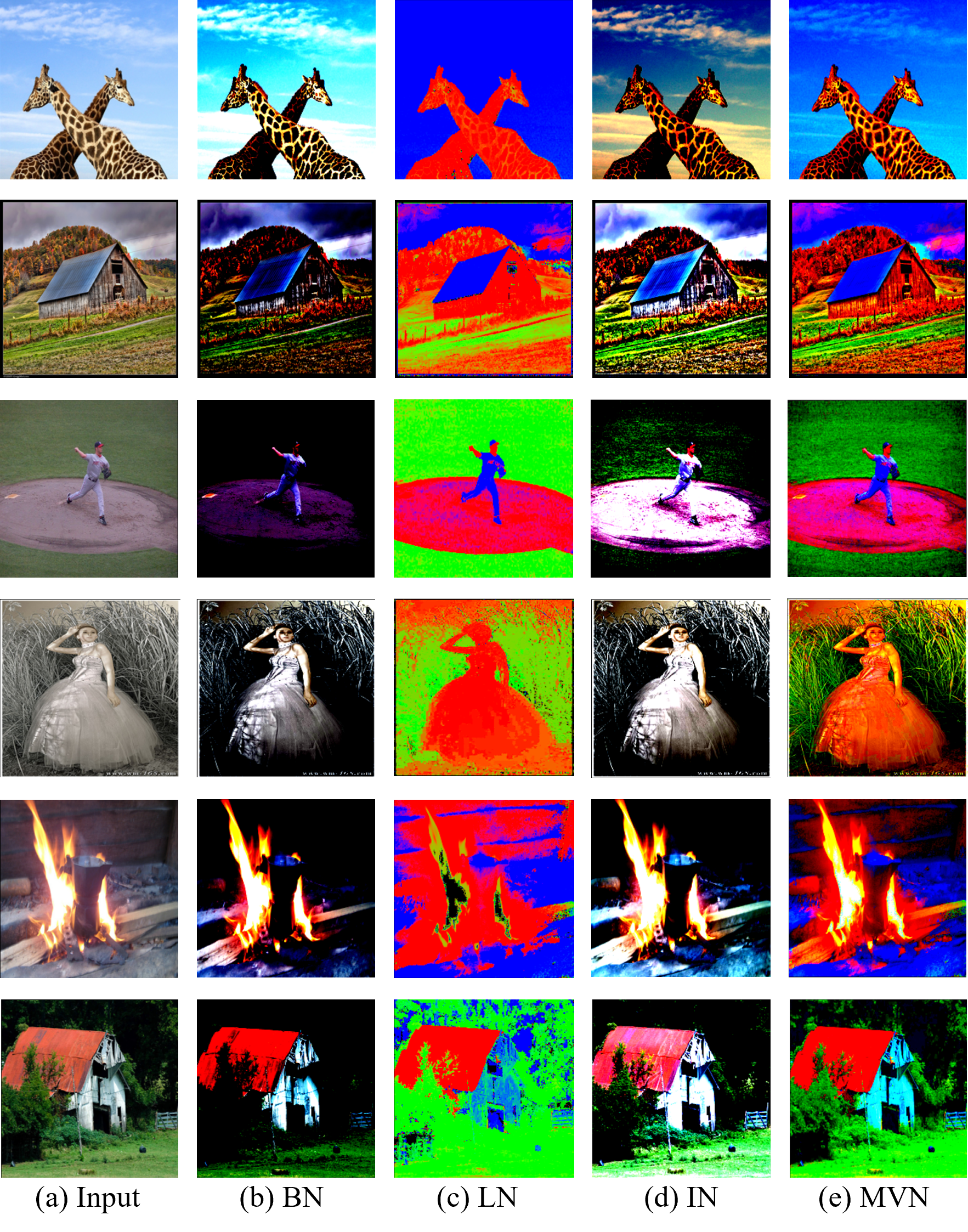}
  % \begin{tabular}{m{36mm}m{32mm}m{30mm}m{40mm}}
  %      (a) Input&(b) BN&(c) LN&(d) BN+LN 
  % \end{tabular}
  \caption{\textbf{Visualization of normalized images from BN, LN, IN, and MVN.}}
  \label{fig:teaser}
\end{figure*}

% As extended from Fig.~\ref{fig:norm_vis},
Fig.~\ref{fig:teaser} shows the visualizations of normalized images from BN, LN, IN, and our MVN.
For the weights of MVN, we apply the ratio of the first block in Stage 1, equal to 0.36, 0.62, and 0.02 for BN, LN, and IN, respectively.
As reflecting the distinct characteristics of three normalized images, through the MVN, we can observe the local pattern-preserved smoothed images.
By this property, MVN is expected to diversify the feature learning.
% BN preserves the spatial detail of the input image, whereas LN oversmoothens it. By taking a straightforward average, the spatially viewed output can intuitively reveal local details as well as global segmentation. 
% We exclude the result of IN since it is the same as that of BN for a single input image.

\section{Pseudo-code in PyTorch}
%Algorithm \ref{alg:code1} and Algorithm \ref{alg:code2} present the PyTorch-like codes for the modules used in the MVN and MVTM.
Algorithm \ref{alg:code1} and Algorithm \ref{alg:code2} are the PyTorch-like pseudo-codes~\cite{paszke2019pytorch} for the MVN and MVTM modules, respectively.
For simplification, we do not consider the channel ordering.

\begin{algorithm*}[hbt!] %htb!
\caption{PyTorch-like Pseudo-code of MVN.}
\label{alg:code1}
\begin{lstlisting}[language=Python]
import torch
import torch.nn as nn

# Multi-View Normalization class
class MultiViewNorm(nn.Module):
    def __init__(self, affine_shape=None):
        super().__init__()
        self.affine_shape = affine_shape
        self.weight = nn.Parameter(torch.ones(self.affine_shape))
        self.bias = nn.Parameter(torch.zeros(self.affine_shape))

        self.ln = nn.LayerNorm(affine_shape, elementwise_affine=False)
        self.bn = nn.BatchNorm(affine_shape, affine=False)
        self.insn = nn.InstanceNorm(affine_shape, affine=False)
        self.norm_weight = nn.Parameter(torch.ones(3,1,1,self.affine_shape))

    def forward(self, x):
        x_ln = self.ln(x)
        x_bn = self.bn(x)
        x_in = self.insn(x)
        
        x = self.norm_weight[0]*x_ln + self.norm_weight[1]*x_bn + self.norm_weight[2]*x_in
        x = x * self.weight + self.bias  #channelwise affine transform
        return x
\end{lstlisting}
\end{algorithm*}

\begin{algorithm*}[hbt!] %hbt!
\caption{PyTorch-like Pseudo-code of MVTM.}
\label{alg:code2}
\begin{lstlisting}[language=Python]
import torch
import torch.nn as nn

# Multi-View Token Mixer class
class MultiViewTokenMixer(nn.Module):
    def __init__(self, in_channels, stage, n_div=4, **kwargs):
        super().__init__()
        self.stage = stage  # 1 2 3 4
        self.split_dim = int(in_channels*2 // n_div)  

        #inverted separable conv
        self.pwconv1 = nn.Linear(in_channels, in_channels*2)  
        self.act1 = StarReLU()
        self.pwconv2 = nn.Linear(in_channels*2, in_channels)

        #stage specificity
        g_kernel, g_pad = ([55, 27, 13, 7], [27, 13, 6, 3])
        self.dim_l = int(self.split_dim * (2 - (self.stage // 2)))  # 50 25 25 0
        self.dim_i = int(self.split_dim * 2)  # 50 50 50 50
        self.dim_g = int(self.split_dim * (self.stage // 2))  # 0 25 25 50

        #multiscale depthwise conv
        self.conv_l = nn.Conv2d(self.dim_l, self.dim_l, groups=self.dim_l, kernel_size=3, padding=1)
        self.conv_i = nn.Conv2d(self.dim_i, self.dim_i, groups=self.dim_i, kernel_size=7, padding=3)
        self.conv_g = nn.Conv2d(self.dim_g, self.dim_g, groups=self.dim_g, 
                                kernel_size=g_kernel[self.stage-1], padding=g_pad[self.stage-1])
                    # decomposed into self.conv_g_h and self.conv_g_v in Stage 1~3
        
    def forward(self, x):
        x = self.pwconv1(x)
        x = self.act1(x)
        
        x_l, x_i, x_g = torch.split(x, (self.dim_l,self.dim_i,self.dim_g), dim=1)
        x_mixed = torch.cat([self.conv_l(x_l), self.conv_i(x_i), self.conv_g(x_g)], dim=1)
            
        x_out = self.pwconv2(x_mixed)
        return x_out

\end{lstlisting}
\end{algorithm*}

% {
%     \small
%     \bibliographystyle{ieeenat_fullname}
%     \bibliography{main}
% }
\end{document}